\documentclass[lettersize,journal]{IEEEtran}
\usepackage{amsmath,amsfonts}
\usepackage{amssymb}
\usepackage{array}
\usepackage{textcomp}
\usepackage{stfloats}
\usepackage{url}
\usepackage{verbatim}
\usepackage{graphicx}
\def\BibTeX{{\rm B\kern-.05em{\sc i\kern-.025em b}\kern-.08em
    T\kern-.1667em\lower.7ex\hbox{E}\kern-.125emX}}
\usepackage{balance}

\usepackage{subcaption}
\usepackage{placeins}
\usepackage{tikz}
\usepackage{algorithm}
\usepackage{algpseudocode}
\makeatletter
\let\NAT@parse\undefined
\makeatother
\usepackage{hyperref}
\usepackage{xcolor}
\usepackage{multirow}
\usepackage{soul} % pour \st CL: package pour barrer du texte (mieux que ulem)
\usepackage{cancel} % CL: package pour barrer un bloc d'équations
\soulregister\cite7 % CL: permet de rendre st compatible avec soul
\soulregister\ref7 % CL : idem
\soulregister\eqref7 % CL idem
\usepackage{comment} % CL: pour commenter des blocs de textes ou eq
\usepackage{booktabs, makecell}

\hypersetup{
    colorlinks=true,        % Enable colored text for links
    linkcolor=nblue,         % Color of internal links (e.g., section links)
    filecolor=magenta,      % Color of file links
    urlcolor=nblue,          % Color of external links
    linkbordercolor=blue,    % Color of the box around internal links
    citecolor=nblue,
    pdfborderstyle={/S/U/W 1} % Option to make the box solid and 1pt wide
}
\definecolor{expander}{HTML}{CC0066}
\definecolor{compactor}{HTML}{3333FF}
\definecolor{support}{HTML}{6C5378}
\definecolor{colorthesis_bleu_fonce}{HTML}{024663}
\definecolor{colorthesis_bleu_clair}{HTML}{006894}
\definecolor{colorthesis_brown}{HTML}{60492C}
\definecolor{colorthesis_green}{HTML}{0C7C59}
\definecolor{colorthesis_red}{HTML}{D64933}
\definecolor{nblue}{rgb}{0,0.263,0.576}
\definecolor{revision}{HTML}{D64933}

\newcommand{\revision}[1]{\textcolor{revision}{#1}}

\newif\ifshowchanges
\showchangesfalse       % clean mode (hide deletions)

\ifshowchanges
    \renewcommand{\revision}[1]{\textcolor{revision}{#1}}
    \newcommand{\deleted}[1]{\st{#1}}
    \newcommand{\deletedfigure}[1]{\xcancel{#1}}
\else
    \renewcommand{\revision}[1]{#1}
    \newcommand{\deleted}[1]{}
    \newcommand{\deletedfigure}[1]{}
\fi

\begin{document}
\bstctlcite{IEEEexample:BSTcontrol} % Nathan
\title{OCSVM-Guided Representation Learning for Unsupervised Anomaly Detection}

\author{Nicolas Pinon, Robin Trombetta and Carole Lartizien
% \thanks{Manuscript received JJ MONTH YYYY; revised JJ MONTH YYYY; accepted JJ MONTH YYYY. Date of publication JJ MONTH YYYY; date of
% current version JJ MONTH YYYY \textit{(Corresponding author: Nicolas Pinon)}}
\thanks{This work has been submitted to IEEE for possible publication. Copyright may be transferred without notice, after which this version may no longer be accessible. \textit{(Corresponding author: Nicolas Pinon)}}
\thanks{Nicolas Pinon, Robin Trombetta and Carole Lartizien are with Univ. Lyon, CNRS UMR 5220, Inserm U1294, INSA Lyon, UCBL, CREATIS, France (e-mails: nicolas.e.pinon@laposte.net; carole.lartizien@creatis.insa-lyon.fr ; robin.trombetta@creatis.insa-lyon.fr )}
% \thanks{Digital Object Identifier: XXXXXXXXXXXXXXX}
}
\markboth{IEEE TRANSACTIONS ON IMAGE PROCESSING}%
{OCSVM-Guided Representation Learning for Unsupervised Anomaly Detection}
\maketitle
% Other titles :
% Task-Aligned Autoencoder for Unsupervised Anomaly Detection
% SVM-Aware Autoencoder for Unsupervised Anomaly Detection
% Autoencoder with SVM-Aligned Representations for Unsupervised Anomaly Detection
% SVM-Guided Autoencoder Training for Unsupervised Anomaly Detection
% SVM-Compatible Representation Learner for Unsupervised Anomaly Detection
% Autoencoder-SVM Coupled Anomaly Detection
% SVM-Oriented Latent Representation Learning for Unsupervised Anomaly Detection
% Unsupervised Anomaly Detection by Joint Optimization of AutoEncoder and One-Class SVM
%\title{Deep representation learning for UAD based on hybrid OCSVM and reconstruction error loss (objective) function}

% HOW MUCH PAGES :
%For the initial submission of a Regular Paper, the manuscript may not exceed 13 double-column pages (using a 10-point font), including title; names of authors and their complete contact information; abstract; text; all images, figures and tables, appendices and proofs; and all references. Supplemental materials and graphical abstracts are not included in the page count. 

% https://signalprocessingsociety.org/publications-resources/information-authors

\begin{abstract}

Unsupervised anomaly detection (UAD) aims to detect anomalies without labeled data, a necessity in many machine learning applications where anomalous samples are rare or not available.
Most state-of-the-art methods fall into two categories: reconstruction-based approaches, which often reconstruct anomalies too well, and decoupled representation learning with density estimators, which can suffer from suboptimal feature spaces.  While some recent methods attempt to couple feature learning and anomaly detection, they often rely on surrogate objectives, restrict kernel choices, or introduce approximations that limit their expressiveness and robustness. To address this challenge, we propose a novel method that couples representation learning with \deleted{an analytically solvable}\revision{the exact} One-Class SVM (OCSVM) \revision{optimization problem}, through a custom loss formulation that directly aligns latent features with the OCSVM decision boundary. The model is evaluated on two tasks: a  benchmark based on MNIST-C, and a challenging brain MRI  lesion detection task. Unlike most methods that focus on large, hyperintense lesions at the image level, our approach succeeds to target small, non-hyperintense lesions, while we evaluate voxel-wise metrics, addressing a more clinically relevant scenario. Both experiments evaluate a form of robustness to domain shifts, including corruption types in MNIST-C and texture or population age variations in MRI.
%Results demonstrate our proposed model's effectiveness,%
Results demonstrate performance and robustness of our proposed model, highlighting its potential for general UAD and real-world medical imaging applications. The source code is available at \url{https://github.com/Nicolas-Pinon/uad_ocsvm_guided_repr_learning}.
\end{abstract}


\section{Introduction}
\label{introduction}

\IEEEPARstart{U}{nsupervised} anomaly detection (UAD) aims to identify patterns in data that deviate significantly from an underlying distribution learned from unlabeled normal samples. It is a critical problem in domains where anomalies are rare, variable, and costly to label, such as fraud detection or medical imaging. In neuroimaging, for instance, detecting subtle or small lesions in MRI scans without annotated anomalies remains an open challenge \cite{lagogiannis2023unsupervised_meissenreview}. Models must not only detect rare and diverse outliers but also generalize reliably to new data distributions, such as those resulting from data acquired on different scanners, or populations with different demographics.
%different scanners, patient demographics, or acquisition noise.

Existing methods fall into two main categories: reconstruction-based approaches and representation learning combined with support or density estimation methods. Autoencoders and their variants are frequently used in reconstruction-based strategies, under the assumption that anomalies will yield higher reconstruction errors. However, these models typically lack structured latent representations, which can lead to high quality reconstruction of never seen anomalies. To overcome this, other methods decouple representation learning from the anomaly scoring process, for instance by training a feature extractor independently from a classifier such as a one-class support vector machine (OCSVM \cite{scholkopf_estimating_2001}). However, this separation can yield to representations not optimized for the decision function computation, leading to suboptimal performance and limited generalization.
Several recent approaches attempt to couple representation learning and anomaly detection more tightly, including methods inspired by Deep SVDD \cite{ruff2018deep}. Yet, these methods often rely on approximations, suffer from hypersphere collapse, or impose strong inductive biases (e.g., linear kernel methods) that limit flexibility and robustness.

To address these limitations, we propose a novel method for UAD that tightly couples an autoencoder-based representation learning with a one-class SVM. Our core contribution lies in a new loss formulation that guides the encoder to produce latent representations optimized for the OCSVM’s boundary. At each training step, the model splits latent samples into two subsets: one to fit the OCSVM boundary and another to enforce that new samples remain within it. This design reduces overfitting to non-relevant features by directly aligning the encoder’s output with the SVM’s discriminative objective. Crucially, it enables the use of \deleted{an exact, analytically solved}\revision{the exact} SVM objective, requiring no approximations or kernel restrictions, thereby preserving the full expressivity of the OCSVM.

To evaluate the proposed method, we conduct two experiments. First, we introduce a new benchmark task based on MNIST-C \cite{MNIST-C_mu2019mnist}, a corrupted version of the MNIST dataset designed to simulate real-world anomalies. This task allows us to rigorously assess the model's performance in a controlled setting and compare it against state-of-the-art UAD methods. Importantly, this experiment evaluates the model's ability to perform anomaly detection under domain shift, as it must generalize across diverse corruption types. Second, we apply the model to two challenging medical imaging tasks: detecting large heterogeneous cancer lesions and detecting subtle (barely visible to the naked eye) brain lesions, both in MRI scans. In medical imaging, many UAD methods have traditionally focused on detecting large, hyperintense lesions, which are often visible and easy to identify, especially through reconstruction-based methods. Our work, in comparison, tackles the problem of detecting lesions that can be small and not necessarily hyperintense, representing a more subtle and clinically significant problem. Additionally, while a significant portion of UAD studies in medical imaging measure performances at the image level, we also assess voxel-wise anomaly detection, thus evaluating precise localization of anomalies within the image. Furthermore, this second series of experiments in neuroimaging evaluates the model's robustness to domain shifts arising from variations in MRI scanners and patient demographics, such as age.

The contributions of this work are twofold:

\begin{itemize}

    \item \textbf{A novel OCSVM-guided representation learning method for general UAD is introduced,  based on a novel loss term aiming at optimizing the representation learner to produce more suitable representations when used in conjunction with OCSVM}

    \item \textbf{We demonstrate the method's applicability on the corrupted MNIST dataset under different domain shifts, as well as on real-world medical imaging tasks, showing improved sensitivity to subtle and non-hyperintense lesions in public brain MRI datasets.}
\end{itemize}

The remainder of this paper is organized as follows: section \ref{sec:biblio} reviews related work on anomaly detection, and then specifically methods used in medical imaging. Section \ref{sec:method} describes our proposed method, detailing the OCSVM-guided representation learning strategy. Sections \ref{sec:1} and \ref{sec:2} present our experimental studies: digit distinction under corruptions using the MNIST-C dataset and lesion detection in brain MRI, respectively. Section \ref{sec:discussion_conclusion} provides a general discussion, including an analysis of the loss components and concludes the paper while outlining potential future research directions.

\section{Related works}
\label{sec:biblio}

Unsupervised Anomaly Detection (UAD) methods can be broadly categorized into three main families: \textit{reconstruction-based methods}, \textit{density estimation-based methods}, and \textit{support estimation-based methods}, as outlined in the review by Ruff et al. \cite{ruff_unifying_2021}. 
All methods share a common objective: modeling the distribution of normal (i.e., non-anomalous) data, often referred to as the normative distribution. Once this distribution is learned, anomalies can be detected as samples that significantly deviate from it.

Representation learning lies at the core of most unsupervised anomaly detection approaches. Reconstruction-based methods use representations as a compression/decompression mechanism, while in practice, both density- and support-based methods typically do not operate directly on raw data, but instead leverage intermediate feature representations, often learned through neural networks, to better capture the structure of normal data.

In the following bibliographic review, we specifically focus on autoencoders due to their simplicity and widespread use as a foundational method for unsupervised feature learning.  While other feature extractors, such as transformer-based models, could also be employed in a similar framework, exploring all possible alternatives is beyond the scope of this work.
We also place a slight emphasis on support estimation methods in order to highlight how our contribution compares to other methods in the same family.

Section~\ref{sec:biblio:ae_recons} covers reconstruction-based methods where the anomaly score is directly derived from the reconstruction error. Section~\ref{sec:biblio:supp_densi_methods}  focuses on support and density estimation methods that use learned representations. We distinguish between decoupled methods, where the representation learning and the anomaly scoring are optimized separately, and coupled methods, \deleted{which jointly optimize both components}\revision{where either both components are optimized jointly, or at least the anomaly detection is guiding the representation learning}, like the method proposed in this work. Finally, section~\ref{sec:biblio:uad_medimage} provides an overview of anomaly detection methods specifically applied to medical imaging.

%dire que l'on va focus sur reconstruction et support car notre méthode utilise un bout de recons mais fait surtout du support.

\subsection{Reconstruction-based methods} 
\label{sec:biblio:ae_recons}

A widely used approach in UAD is to leverage an autoencoder's (AE) ability to reconstruct normal data while failing to accurately reconstruct anomalies. As detailed in the review by Ruff et al. \cite{ruff_unifying_2021}, reconstruction-based methods assume that, after training on normal samples, an autoencoder will learn a compressed representation that captures essential features of the normal data distribution. When presented with an anomalous input, the reconstruction error is expected to be significantly higher due to the model's inability to generalize to unseen, out-of-distribution patterns.

Early approaches relied on simple autoencoders trained with standard mean squared error or cross-entropy loss, where anomalies were detected based on high reconstruction error \cite{kramer1992autoassociative}. This paradigm has been widely applied to image anomaly detection \cite{sakurada2014anomaly, beggel2020robust, pinon2024unsupervised_sec3.1} and extended to various domains, such as industrial defect detection \cite{bergmann2019mvtec} or medical images \cite{baur2021autoencoders}. Variational Autoencoders (VAE) introduced a probabilistic constraint on the latent space, which helps regularize representations, but they often struggle to clearly separate normal from anomalous reconstructions due to their tendency to generate blurry outputs \cite{zimmerer2019high}.

Hybrid methods, known as restoration methods have emerged, which combine the reconstruction error with an estimation of the density of the distribution of normal samples in the autoencoder's latent space. These methods aim to ``heal" the image by restoring it to the normal distribution (thus erasing the anomaly) and then comparing it to the original image through the reconstruction error. One example is the work by Wang et al. \cite{wang_image_2020}, which applies this approach to industrial images by using a quantized autoencoder (VQ-VAE) in conjunction with an autoregressive model (PixelSnail \cite{chen2018pixelsnail}) for density estimation in the latent space. Another type of restoration methods has gained recent popularity for anomaly detection in images: diffusion models, where the image is first partially noised, and then denoised with a UNet-like model, effectively providing a restored image \cite{he2024diffusion}.

Another alternative direction involves synthetic anomaly detection (also called self-supervised learning strategies \cite{golan2018deep, CAI2025_medianomaly}), where synthetic anomalies are added to the data during training of a supervised method. This approach, also proved effective in medical imaging \cite{kascenas2022anomaly, tan2022detecting}, suffers from a severe drawback : the synthetic anomalies distribution must match the (unknown) true anomaly distribution, therefore imposing a strong prior on anomalies that can be detected.

Despite their effectiveness, Ruff et al. \cite{ruff_unifying_2021} highlight several limitations of reconstruction-based methods. Autoencoders may generalize too well, inadvertently reconstructing anomalies with low error, which weakens their discriminative power \cite{zhang2021understanding}. Also, reconstruction error alone does not explicitly define a geometrically-coherent decision boundary between normal and anomalous data, making it hard to calibrate anomaly scores. These challenges motivate alternative approaches where autoencoders serve as representation learners rather than direct anomaly detectors, as discussed in section~\ref{sec:biblio:supp_densi_methods}.

\subsection{Support/density estimation methods}
\label{sec:biblio:supp_densi_methods}

Density and support estimation methods attempt to explicitly characterize the distribution of normal data either by modeling its density or by learning a decision boundary that encloses the normal data.  They typically rely on representation learning techniques to effectively model the structure of normal data  that can be coupled with classical methods like One-Class SVM (OCSVM \cite{scholkopf_estimating_2001}), Support Vector Data Description (SVDD \cite{tax_support_2004} and their variants, Gaussian Mixture models, etc.

In this section, we distinguish between decoupled methods, where the representation learner is trained separately (\ref{sec:biblio:supp_densi_methods:decoupled}) before applying a support or density estimation method, and coupled methods (\ref{sec:biblio:supp_densi_methods:coupled}), where the representation learning process is influenced by the anomaly detection objective.

\subsubsection{Decoupled methods}
\label{sec:biblio:supp_densi_methods:decoupled}

A common approach is to first train an autoencoder to reconstruct its input, thus providing an encoder capable of producing a compressed representation of the input and then apply a separate anomaly detection method on the learned latent representations such as multivariate Gaussian in PaDiM \cite{defard2021padim}, clustering in Perera and Patel \cite{perera2019learning} or OCSVM in \cite{mabu2021anomaly, beggel2020robust, pinon2024unsupervised_sec3.1} .

%Another approach is to  train a OCSVM-like model on the autoencoder-based representation to learn a decision boundary enclosing the normative distribution. 
This OCSVM-based approach was applied to industrial images \cite{pinon2024unsupervised_sec3.1} and synthetic aperture radar images \cite{mabu2021anomaly}. In both cases, a convolutional autoencoder is trained on normal samples, and the encoder's latent features are fed to an OCSVM for anomaly detection. In \cite{mabu2021anomaly}, the features are further reduced via PCA, and as in \cite{beggel2020robust} a discriminator is used.

Decoupled methods often suffer from a sub-optimal alignment between the learned representations and the anomaly detection objective.  This mismatch can lead to degraded performance, particularly in complex or high-dimensional settings where anomaly structures are subtle. 

\subsubsection{Coupled methods}
\label{sec:biblio:supp_densi_methods:coupled}

Coupled methods aim to address this limitation by \deleted{integrating the representation learning and support/density estimation steps into a unified framework}\revision{either jointly optimizing the representation learning and support/density estimation steps or by guiding the representation learning with the anomaly detection objective}, thereby encouraging the latent space to be more directly optimized for the detection task. A foundational example is Deep SVDD (DSVDD \cite{ruff2018deep}), which replaces the implicit dual space mapping of traditional SVDD by an explicit modeling (thus approximated) with a neural network. The normal data points are projected in a dual space where they must fit into an hypersphere of learned radius (soft-variant) or just compacted around a predefined center (hard-variant). Anomalies are then identified by measuring the distance to the center (hard) or to the hypersphere (soft). The method is evaluated on several standard image datasets, including MNIST and CIFAR-10, where it demonstrates better performance than kernel-based baselines such as OCSVM.

Different approaches, namely DSPSVDD \cite{zhang_anomaly_2021}, DVAESVDD \cite{zhou2021vae}, DASVDD \cite{hojjati2023dasvdd}, CDSVDD \cite{xing2023contrastive} or Patch SVDD \cite{yi2020patch}, are built on the seminal Deep SVDD formulation to both address the hypersphere collapse issue and attempt to improve the discriminative power of the learned representation. They all were shown to outperform it based on datasets such as MNIST, Fashion-MNIST \cite{xiao2017fashion} or MVTecAD \cite{bergmann2019mvtec}.

Deep Structure Preservation SVDD (DSPSVDD) \cite{zhang_anomaly_2021} enhances Deep SVDD by first pre-training an autoencoder and then adding the deep SVDD term in the loss for further fine-tuning. The major difference is that the reconstruction loss term is still present in the fine-tuning. 

In a similar fashion, VAE-based Deep SVDD (DVAESVDD \cite{zhou2021vae}) optimizes a VAE's reconstruction loss and the SVDD's hypersphere loss. Contrastive Deep SVDD (CDSVDD \cite{xing2023contrastive}) leverages contrastive learning to improve the discriminative power of the learned representations by minimizing both the contrastive loss and the SVDD loss.

 % c is initialized as 0 and then recomputed as the mean of points at each epoch (as in VAESVDD), and also proportion of points K (=0.9) used for recons and other for c. weight param is fixed with a certain procedure.

Beyond SVDD-based formulations, Zong et al. \cite{zong_deep_2018} introduce the Deep Autoencoding Gaussian Mixture Model, which combines a compression network with a GMM applied in the latent space. The loss function integrates the reconstruction error, the GMM log-likelihood, and a regularization term; it was originally tested on tabular datasets (KDDCup99, Thyroid, Arrhythmia).

\revision{Closer to our work, AE-1SVM \cite{nguyen2019scalable} jointly trains an autoencoder and an OCSVM by minimizing the primal hinge-loss objective with SGD, the RBF kernel being approximated with random Fourier features.}

Other coupled methods include one-class GAN (OCGAN) \cite{perera2019ocgan}, which uses adversarial training to enforce that every normal samples are distributed as a uniform distribution and that every interpolated sample from this distribution output a normal-looking image. 

Overall, in coupled methods, \revision{the representation learning} seems to benefit from \deleted{end-to-end optimization}\revision{the guidance providied by the anomaly detection process}\deleted{, where the representation learning and anomaly detection objectives are jointly optimized}, as this could ensure that the learned features are directly tailored for anomaly discrimination.  While coupled approaches seem to surpass their decoupled counterpart in the cited studies, the diversity of evaluation protocols and datasets makes generalization of conclusions difficult. To the best of our knowledge, no comprehensive study has been conducted to systematically assess the benefits of coupling representation learning with anomaly detection, compared to decoupled approaches. Also, to this day, no method makes use of the full flexibility offered by the kernel-representation of OCSVM or SVDD: all methods use approximations or limitations regarding the type of kernel used for dual space mapping.

Moreover, most existing studies focus on standard, low-complexity datasets such as MNIST, Fashion-MNIST, or CIFAR-10, which do not reflect the challenges of real-world applications. In particular, the medical imaging domain, despite its complexity and practical importance, remains largely unexplored in this context. This highlights the need for a dedicated review of UAD methods in medical imaging, which we present in section \ref{sec:biblio:uad_medimage}.

\subsection{Unsupervised anomaly detection for medical images}
\label{sec:biblio:uad_medimage}

%%%%%%%%%%%%%%%%%%%%%%%%%%%%%%%%%%%%%%%%%%%%%%%%%%%%%%%%%%%%
%%%%%% RESSOURCES SI JAMAIS IL MANQUE DES CHOSES : %%%%%%%
%%%%%%%%%%%%%%%%%%%%%%%%%%%%%%%%%%%%%%%%%%%%%%%%%%%%%%%%%%%%
% https://arxiv.org/pdf/2012.02364 + tschunig + medianomaly
%Dans \cite{mini_review_tschuchnig2021anomaly} il y a un tableau spécifique à brain MRI
%\red{46 à 60 dans Ruff review}
%\red{parler de la diffusion ? comme on en fait pas}

In this section, we focus on Unsupervised Anomaly Detection (UAD) methods specifically applied to medical imaging. While the broader field of medical anomaly detection encompasses a wide range of modalities and anatomical regions, we restrict our discussion to studies that align with our focus on brain MRI, based on recent reviews and benchmark analyses in the domain~\cite{CAI2025_medianomaly, Behrendt_UAD_review_MEDIA26}.

%Reconstruction-based methods presented in section \ref{sec:biblio:ae_recons} have been widely applied to medical imaging. Baur et al. \cite{baur2021autoencoders} conducted a comprehensive comparative benchmark of various autoencoder architectures, including classical, variational and adversarial autoencoders for detecting large tumor lesions (Gliomas) as well as hyperintense smaller lesions in brain MRI datasets such as MSLUB \cite{MSLUB_lesjak2018novel} and MSSEG \cite{MSSEG2015_commowick2018objective}. Their findings highlight the effectiveness of reconstruction-based approaches for identifying small, hyperintense lesions, which are common in conditions like multiple sclerosis. The same authors  proposed a UNet-like autoencoder architecture \cite{baur_modeling_2021} demonstrating strong performance for detecting small hyperintense lesions on the WMH challenge dataset.

Reconstruction-based methods presented in section \ref{sec:biblio:ae_recons}, have been widely applied to brain lesion detection in MRI.  Baur et al. \cite{baur2021autoencoders} conducted a comprehensive comparative benchmark of various autoencoder architectures, including classical, variational and adversarial autoencoders for detecting large tumor lesions (Gliomas) as well as hyperintense lesions in brain MRI datasets such as MSLUB \cite{MSLUB_lesjak2018novel} and MSSEG \cite{MSSEG2015_commowick2018objective}. Their findings highlight the effectiveness of reconstruction-based approaches for identifying small, hyperintense lesions, which are common in conditions like multiple sclerosis. The same authors  proposed a UNet-like autoencoder architecture \cite{baur_modeling_2021} demonstrating strong performance for detecting small hyperintense lesions on the WMH challenge dataset. 

Pinaya et al. \cite{pinaya_unsupervised_2022} introduced a restoration-based approach using a Vector Quantized Variational Autoencoder (VQ-VAE) coupled with a transformer model for density estimation in the latent space. This method was evaluated on multiple neuroimaging datasets, including MSLUB, BraTS, and WMH, further highlighting the utility of reconstruction-based techniques for hyperintense lesion detection. Additionally, Ramirez et al. \cite{ramirez_deep_2020} used VAEs to detect anomalies in Parkinson's patients' brain MRI, while Zimmerer et al. \cite{zimmerer2019unsupervised} and Zhao et al. \cite{zhao2022ae} employed VAEs for brain tumor segmentation, leveraging reconstruction errors as anomaly scores.

Diffusion models have been recently proposed as an alternative to auto-encoder based architectures in UAD for brain lesions detection in MRI. As for AE-based models, the underlying assumption is that these models will generate anomaly-free images when inputted pathological images at inference. The anomaly score map is then computed based on scoring functions accounting for the standard reconstruction error map. Like autoencoders, these models suffer from the need to compromise the sensitivity to anomalous regions, increasing with high corruption (noise for diffusion models and compression for AE), and the fidelity of the reconstructed normal regions (decreasing with noise/compression), as seen in \cite{liang2025itermask3d}. Several architectures have been proposed to enhance detection performance of these models. Some authors proposed employing more elaborate noise types, e.g. multiscale or simplex noise in DAE \cite{kascenas2023role} and AnoDDPM \cite{wyatt_anoddpm_2022}) instead of standard Gaussian noise. Very recent contributions leverage information from the original image, either by conditioning the denoising process to the original images, like cDDPM \cite{behrendt2024mhd, Behrendt_cDDPM_CIBM25} or by masking strategies, to focus the denoising process on the lesion area so as to preserve the details of regions predicted as healthy, like AutoDDPM \cite{bercea2023mask} and THOR \cite{Bercea_THOR_MICCAI24} which apply implicit guidance only during inference or MAD-AD \cite{Beizaee_MADAD_IPMI25} which incorporates information from the original image during both training and inference.

In addition to reconstruction-based methods or methods based on synthetic anomaly generation, support/density estimation approaches,  presented in section~\ref{sec:biblio:supp_densi_methods}, have also been applied to medical imaging. For example, we proposed  to employ autoencoders as feature extractors, followed by OCSVM for anomaly detection \cite{alaverdyan_regularized_2020, pinon_one-class_2023, pinon_brain_2023}. In \cite{alaverdyan_regularized_2020}, we utilized a localized OCSVM approach to detect challenging epileptogenic lesions in a private dataset, while in \cite{pinon_one-class_2023} and \cite{pinon_brain_2023}, we proposed a patient-specific OCSVM framework evaluated on the WMH dataset in the former and on the public PPMI dataset (related to Parkinson disease) in the later. Furthermore, Azami et al. \cite{azami_plos_one_2016detection} and Bowles et al. \cite{bowles_brain_2017} used OCSVM for brain MRI anomaly detection, the latter applying it to unsupervised brain lesion segmentation by modeling white and gray matter voxels. 

Cai et al. \cite{CAI2025_medianomaly} performed a wide benchmark on image-level anomaly detection on medical imaging datasets, and image-level and voxel-level anomaly detection specifically for brain MRI on different datasets including the BraTS dataset \cite{brats2014} depicting brain tumors but not any dataset depicting smaller brain lesions like WMH or MSLUB. While this benchmark does not evaluate support/density estimation methods on the voxel-level anomaly detection task, they evaluate a wide variety of methods based on reconstruction error including DAE \cite{kascenas2022denoising}, AnoDDPM \cite{wyatt_anoddpm_2022} and AutoDDPM \cite{bercea2023mask}  or models based on synthetic anomaly generation. They find that reconstruction-based methods outperform other methods for voxel-level anomaly detection. On the BraTS dataset, DAE \cite{kascenas2022denoising} is shown to outperform autoDDPM \cite{bercea2023mask}, itself outperforming anoDDPM with simplex noise \cite{wyatt_anoddpm_2022}.

As pointed out in the literature \cite{Behrendt_UAD_review_MEDIA26}, recent benchmark studies dedicated to the detection and localization of small lesions (e.g. MSLUB or WMH) are lacking. Results compiled from the 2021 benchmark study of Baur et al \cite{baur2021autoencoders} for autoencoder based models and from recent papers including state-of-the art diffusion models (\cite{Beizaee_MADAD_IPMI25, behrendt2024mhd}) highlight that cDDPM \cite{behrendt2024mhd, Behrendt_cDDPM_CIBM25} combined with an anomaly scoring function based on the Mahalanobis distance outperformed anoDDPM \cite{wyatt_anoddpm_2022}, which itself performed better than DAE \cite{kascenas2022denoising} for the detection and localisation of WMH and MSLUB lesions. In two recent studies of Beizaee et al. \cite{Beizaee_MADAD_IPMI25, Beizaee_Reflect_MICCAI2025}, anoDDPM \cite{wyatt_anoddpm_2022} was also shown to outperform autoDDPM \cite{bercea2023mask}, itself performing better than DAE \cite{kascenas2022denoising} on small stroke lesions of the ATLAS dataset.

The recent review of Behrendt et al \cite{Behrendt_UAD_review_MEDIA26} pointed out another issue in the evaluation of medical anomaly detection methods in brain MRI related to the predominance of hyperintense lesions in benchmark datasets. As first noted by Meissen et al. \cite{meissen2021pitfalls}, many state-of-the-art methods are evaluated on anomalies that are significantly brighter than the surrounding tissue in the MRI image (e.g. FLAIR), such as those in the BraTS and WMH datasets. This raises concerns about the generalizability of these methods to more challenging anomalies, such as those with subtle intensity differences or complex morphological characteristics. Some groups including ours (\cite{meissen2021challenging, Pinon_Thesis_2024}) demonstrated that simply thresholding these MRI images could achieve competitive performance on hyperintense lesion detection e.g. WMH lesion in FLAIR imaging, highlighting the need for more rigorous evaluation protocols and diverse datasets.

Despite encouraging results on hyperintense lesions, the performance of unsupervised anomaly detection methods on more challenging, publicly available medical imaging datasets remains largely unevaluated. Autoencoder-based reconstruction methods continue to serve as strong baselines. In contrast, support and density estimation approaches (decoupled \ref{sec:biblio:supp_densi_methods:decoupled}) remain underexplored in this context, often evaluated only on private datasets or omitted from comparative benchmarks. Also, to the best of our knowledge, no coupled (\ref{sec:biblio:supp_densi_methods:coupled}) method that \deleted{jointly optimizes feature representation and anomaly detection}\revision{guides the representation learning with the anomaly detection process} has been applied to medical imaging.

\section{Method: OCSVM-guided representation learning}
\label{sec:method}

\begin{figure*}
    \centering
    \includegraphics[width=0.9\linewidth]{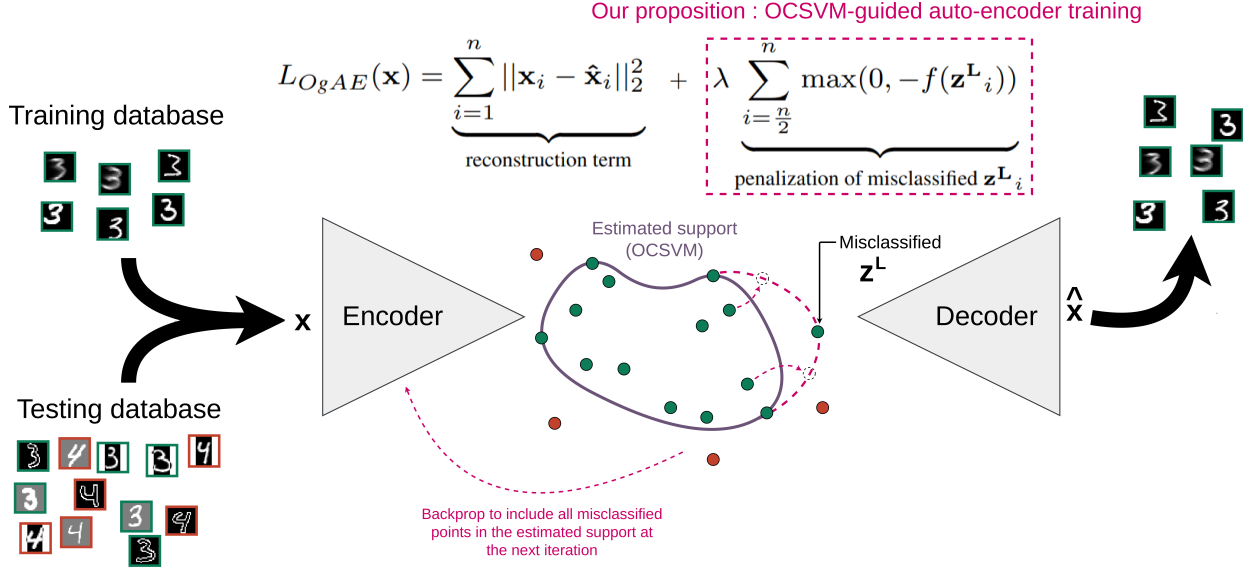}
    \caption{Graphical abstract of the proposed method. During training, the autoencoder must both minimize the reconstruction error between input and output and a new loss term (section \ref{sec:method:coupling}) that guides the encoder towards representations that are more fitted for support estimation with OCSVM.}
    \label{fig:graph_abstract}
\end{figure*}

The method we propose is presented in Figure \ref{fig:graph_abstract}. An autoencoder is used for representation learning, while a OCSVM is used to estimate the normal data distribution support. The main goal of the term that we add to the loss function of the autoencoder is to use normal samples that are misclassified during training (projected outside the support) to modify the representation space such that these misclassified samples will be included in the estimated support at the next iteration.
As we stated in section \ref{sec:biblio:supp_densi_methods}, the proposed architecture is compatible with any representation learner (including transformers). 
Section \ref{sec:method:decoupled_lr_and_ad} details the main idea of the method without coupling by describing the representation learning step \ref{sec:method:decoupled_lr_and_ad:ae_repr}, followed by the anomaly detection step \ref{sec:method:decoupled_lr_and_ad:ocsvm}. Section \ref{sec:method:coupling} describes our contribution: coupling of the two steps through the OCSVM-guidance of the representation learning.

\subsection{Decoupled representation learning and anomaly detection}
\label{sec:method:decoupled_lr_and_ad}

\subsubsection{Representation learning with autoencoder}
\label{sec:method:decoupled_lr_and_ad:ae_repr}

%To train the encoder to perform dimensionality reduction, 
To learn efficient and compressed representations, we train the autoencoder to reconstruct as accurately as possible an input batch of normal samples \footnote{Input batches in Experiment \textcolor{nblue}{\hyperref[sec:1]{1}} will be batches of whole images and in Experiment \textcolor{nblue}{\hyperref[sec:2]{2}} batches of image patches, but this method can be used with any type of data (even non-image, if the autoencoder is adapted).} $(\mathbf{x}_1, \dots, \mathbf{x}_n)$, based on the classical MSE loss: 

\footnotesize
\begin{equation} L_{AE}(\mathbf{x}_1, \dots, \mathbf{x}_n) = \sum_{i=1}^n ||\mathbf{x}_i-\mathbf{\hat{x}}_i||_2^2\end{equation}
\normalsize

Where $\mathbf{\hat{x}}_i$ is the reconstruction of $\mathbf{x}_i$. After training, the decoder is discarded and the encoder is used, frozen, to perform dimensionality reduction of samples $\mathbf{x}$ into their latent representation $\mathbf{z}$.

\subsubsection{Anomaly detection with one-class SVM}
\label{sec:method:decoupled_lr_and_ad:ocsvm}

To perform the detection of anomalies, we estimate the support of the normal data (the boundaries of the normative distribution) with a One-Class SVM (OCSVM \cite{scholkopf_estimating_2001}). This is done by constructing a decision function $f$, positive on the estimated support of the distribution of normal samples $\mathbf{z}_i$, negative elsewhere and null on the frontier. The normal samples are first mapped to a high dimensional space by a feature map $\mathbf{\Phi}(\cdot)$ associated with a kernel $k$ such that $k(\mathbf{z}_i,\mathbf{z}_j)$ = $\mathbf{\Phi}(\mathbf{z}_i) \cdot \mathbf{\Phi}(\mathbf{z}_j)$. As the problem is linear in this re-description space, the parameters $\mathbf{w}$ and $\rho$ of the hyperplane $\mathbf{w} \cdot \mathbf{\Phi}(\mathbf{z}) - \rho = 0$ are obtained by solving a convex optimization problem, presented in equation \ref{eq:ocsvm_z}, aiming at maximizing the distance of the hyperplane from the origin.

\footnotesize\begin{equation}
\begin{aligned}
& \underset{\mathbf{w}, \rho, \boldsymbol{\xi}}{\text{min} }
& &  \frac{1}{2}||\mathbf{w}||^2 + \frac{1}{\nu n} \sum_{i=1}^{n} \xi_i - \rho  \\
& \text{subject to}
& &  \langle \mathbf{w}, \mathbf{\Phi}(\mathbf{z}_i) \rangle \geq \rho - \xi_i & i \in [1, n] \\        
& & &   \xi_i \geq 0 & i \in [1, n] 
\end{aligned}
\label{eq:ocsvm_z}
\end{equation}\normalsize

The decision function can then be expressed as $f(\mathbf{z}) = \mathbf{w}^* \cdot \mathbf{\Phi}(\mathbf{z}) -\rho^*$, with $\mathbf{w}^*$ and $\rho^*$ the solutions of the optimization problem.

Through a process known as the kernel trick, the problem is actually solved in its dual form leading to the following expression of the decision function

\footnotesize\begin{equation} f(\mathbf{z}) = \sum_{j=1}^{n} \alpha_j^* k(\mathbf{z}_j, \mathbf{z}) - \rho^* 
\label{eq:ocsvm_decision_function}\end{equation}\normalsize

which corresponds to a weighted mean of the kernel distance to each normal samples $k(\mathbf{z}_j, \mathbf{z})$. The Lagrange multipliers of the dual problem ($\alpha_j^*$) are sparse and $\rho^*$ is derived from them.

At inference, to obtain the anomaly score of a new sample $\mathbf{x}$, it must first go through the encoder to obtain its latent representation $\mathbf{z}$, and then through the decision function $f$. Note that this score will be positive if the sample is within the distribution and negative if outside. The more negative the score, the further the sample is from the normal distribution and thus the more suspicious it will be considered.

% Arguments méthode :
    % Without expander: Normal data collapses into a tiny cluster (poor generalization).
    %Without the expander, the compactor alone could collapse the boundary (e.g., all points mapped to a single value). Cite Deep SVDD’s caveats (Ruff et al., 2018) on representation collapse.
    % Without compactor : boundary gets endlessly expanded and might include anomalies
    % With expander and compactor: Normal data is tight but leaves "breathing room" for the boundary to exclude outliers.

    %While our method never observes anomalies during training, the OCSVM’s hypersphere boundary is optimized to compactly describe the manifold of normal data in latent space. By jointly training the encoder to align with this boundary (whether by expanding it or comapcting it), we implicitly encourage deviations from this manifold to correspond to anomalous behavior. This follows the key assumption of UAD: anomalies are outliers to the normal data’s distribution, and thus need not be explicitly modeled.

\subsection{Coupling: OCSVM-guidance of the representation learning}
\label{sec:method:coupling}

\begin{figure}
    \centering
    \includegraphics[width=0.85\columnwidth]{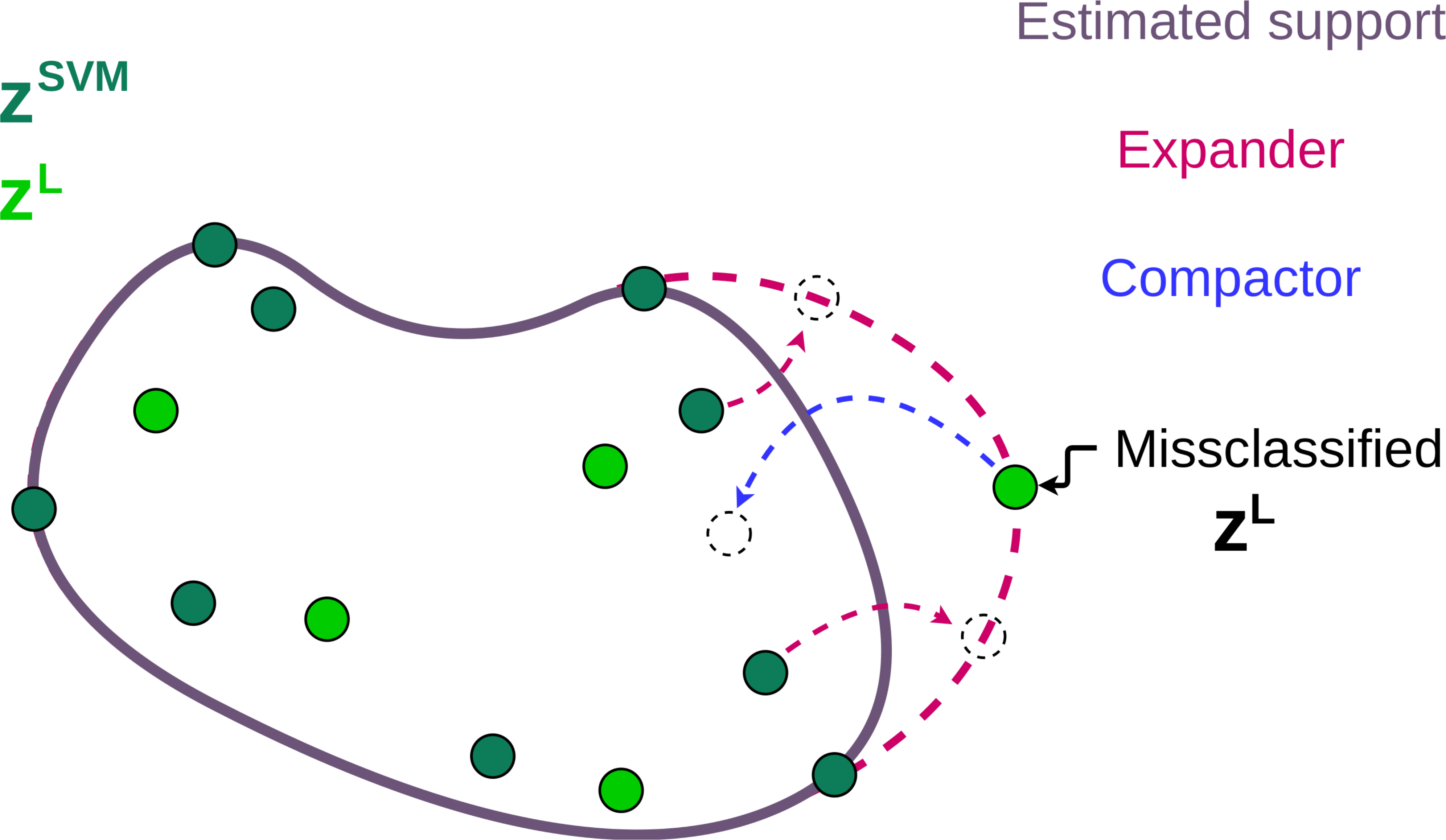}
    \caption{Visualization of the two terms present in the proposed loss : both terms are based on the idea to use the misclassified $\mathbf{z^{\scriptscriptstyle L}}$ to steer the representations towards SVM-compatible features. While the \textcolor{expander}{expander} term focus on moving the $\mathbf{z^{\scriptscriptstyle SVM}}$ to expand the \textcolor{support}{estimated support}, the \textcolor{compactor}{compactor} term focus on moving the $\mathbf{z^{\scriptscriptstyle L}}$ inside the \textcolor{support}{estimated support}.}
    \label{fig:expander_jz}
\end{figure}

We describe in this section our contribution: a novel OCSVM-guidance (O\textit{g}) loss term. The goal of this loss is to align as best as possible the representation of the encoder with the downstream task of estimating the support of the normal distribution with the OCSVM. This is performed by splitting each training batch into two, one used for support estimation ($\mathbf{z^{\scriptscriptstyle SVM}}$) and one for loss computation ($\mathbf{z^{\scriptscriptstyle L}}$).

The OCSVM-guidance is divided into two terms: the \textcolor{expander}{expander} and the \textcolor{compactor}{compactor}, as represented in Figure \ref{fig:expander_jz}. The \textcolor{compactor}{compactor} term makes the estimated support more compact by \textit{moving misclassified normal training samples} inside the estimated support: this ensures the support stays compact and allows anomalies to fall outside the support. To prevent collapsing of the support, as can happen in deep SVDD, the \textcolor{expander}{expander} term \textit{moves the boundary} such that misclassified normal training samples fall inside the estimated support. By training the encoder to align with the estimated support (whether by expanding it or compacting it), we implicitly encourage deviations from this manifold to correspond to anomalous behavior, thus learning OCSVM-compatible features while avoiding irrelevant overfitting.

As stated, one batch of samples, after encoding, $\mathbf{z} = (\mathbf{z}_1, \dots, \mathbf{z}_n)$ is split into two: the part used to solve the OCSVM problem ($\mathbf{z^{\scriptscriptstyle SVM}}$) and the other for the loss computation ($\mathbf{z^{\scriptscriptstyle L}}$):

\footnotesize
\begin{equation}
\mathbf{z}_i = 
\begin{cases}
    \mathbf{z}^{\scriptscriptstyle SVM}_i & \text{for } 1 \leq i \leq \frac{n}{2}, \\
    \mathbf{z}^{\scriptscriptstyle L}_i    & \text{for } \frac{n}{2} < i \leq n.
\end{cases}
\end{equation}
\normalsize

At each batch, we solve the OCSVM problem for the $\mathbf{z^{\scriptscriptstyle SVM}}_i$, which will give the optimal $\boldsymbol{\alpha}$ and $\rho$: $\boldsymbol{\alpha}^*$ and $\rho^*$.

The proposed $L_{O\textit{g}AE}$ loss is composed of a standard reconstruction error term and the OCSVM-guidance (O\textit{g}) term, which penalizes the misclassified $\mathbf{z^{\scriptscriptstyle L}}_i$ (that are not used to compute the SVM problem):

\footnotesize\begin{equation}  L_{O\textit{g}AE}(\mathbf{x}) = \underbrace{\sum_{i=1}^n ||\mathbf{x}_i-\mathbf{\hat{x}}_i||_2^2}_{\text{reconstruction term}}  + \lambda \underbrace{\sum_{i=\frac{n}{2}}^n \max(0, -f(\mathbf{z^{\scriptscriptstyle L}}_i))}_{\text{penalization of misclassified } \mathbf{z^{\scriptscriptstyle L}}_i}  \end{equation}\normalsize

The second term, weighted by $\lambda$, indeed penalizes only the misclassified $\mathbf{z^{\scriptscriptstyle L}}_i$, as the decision function outputs positive values for correctly classified $\mathbf{z^{\scriptscriptstyle L}}_i$, and thus $\max(0, -f(\mathbf{z^{\scriptscriptstyle L}}_i))$ is 0. Misclassified $\mathbf{z^{\scriptscriptstyle L}}_i$ are penalized proportionally to their euclidean distance to the estimated hyperplane.

The interest of separating the latent representation vectors into two parts $\mathbf{z^{\scriptscriptstyle SVM}}$ and $\mathbf{z^{\scriptscriptstyle L}}$ appears here: as the SVM frontier is estimated on the $\mathbf{z^{\scriptscriptstyle SVM}}_i$, most of them are correctly classified. This justifies the use of another set of latent vectors $\mathbf{z^{\scriptscriptstyle L}}$. Penalizing samples not used for the support estimation could also be viewed as a way to penalize bad generalization to unseen samples. We can develop $L_{O\textit{g}AE}$ with the expression of $f$ from equation \ref{eq:ocsvm_decision_function}:

\footnotesize
\begin{equation}
    \begin{split}
        L_{O\textit{g}AE}(\mathbf{x}) &= \sum_{i=1}^n ||\mathbf{x}_i-\mathbf{\hat{x}}_i||_2^2 \\
        &\quad +\lambda \sum_{i=\frac{n}{2}}^n \max\left(0, - \sum_{j=1}^{\frac{n}{2}} \alpha_j^* k(\mathbf{z^{\scriptscriptstyle SVM}}_j, \mathbf{z^{\scriptscriptstyle L}}_i) - \rho^* \right)
    \end{split}
\end{equation}
\normalsize

Recall that $\boldsymbol{\alpha}^*$ and $\rho^*$ are functions of the $\mathbf{z^{\scriptscriptstyle SVM}}_i$. If we separate the $\lambda$-term into what depends on the $\mathbf{z^{\scriptscriptstyle SVM}}_i$ and what depends on the $\mathbf{z^{\scriptscriptstyle L}}_i$, using the stopgradient operator $\mathrm{sg}[.]$ and two linked coefficients $\beta_1 + \beta_2 = 1$, we can write $L_{O\textit{g}AE}$ as:

% \footnotesize\begin{equation}
% \begin{aligned}
% L_{O\textit{g}AE}(\mathbf{x}) = & 
% \sum_{i=1}^n ||\mathbf{x}_i-\mathbf{\hat{x}}_i||_2^2  \\
% & + \lambda  \beta_1\overbrace{\sum_{i=\frac{n}{2}}^n \max(0, - \sum_{j=1}^{\frac{n}{2}} \alpha_j^* k(\mathbf{z^{\scriptscriptstyle SVM}}_j, \mathrm{sg}[\mathbf{z^{\scriptscriptstyle L}}_i]) - \rho^* )}^{\text{\textcolor{expander}{Gradient flow only through the }$\mathbf{z^{\scriptscriptstyle SVM}}_i$}} \\
% & + \lambda  \beta_2\underbrace{\sum_{i=\frac{n}{2}}^n \max(0, - \sum_{j=1}^{\frac{n}{2}} \mathrm{sg}[\alpha_j^*] k(\mathbf{\mathrm{sg}[z^{\scriptscriptstyle SVM}}_j], \mathbf{z^{\scriptscriptstyle L}}_i) - \mathrm{sg}[\rho^*] )}_{\text{\textcolor{compactor}{Gradient flow only through the }$\mathbf{z^{\scriptscriptstyle L}}_i$}}
% \end{aligned}
% \label{eq:jz_final}
% \end{equation}\normalsize

\footnotesize
\begin{flushleft}
\begin{equation}
\begin{array}{l}
L_{O\textit{g}AE}(\mathbf{x}) = 
\sum_{i=1}^n ||\mathbf{x}_i-\mathbf{\hat{x}}_i||_2^2  \\[0.8em]
+ \lambda  \beta_1\overbrace{\sum_{i=\frac{n}{2}}^n \max(0, - \sum_{j=1}^{\frac{n}{2}} \alpha_j^* k(\mathbf{z^{\scriptscriptstyle SVM}}_j, \mathrm{sg}[\mathbf{z^{\scriptscriptstyle L}}_i]) - \rho^* )}^{\text{\textcolor{expander}{Gradient flow only through the }$\mathbf{z^{\scriptscriptstyle SVM}}_i$}} \\
+ \lambda  \beta_2\underbrace{\sum_{i=\frac{n}{2}}^n \max(0, - \sum_{j=1}^{\frac{n}{2}} \mathrm{sg}[\alpha_j^*] k(\mathbf{\mathrm{sg}[z^{\scriptscriptstyle SVM}}_j], \mathbf{z^{\scriptscriptstyle L}}_i) - \mathrm{sg}[\rho^*] )}_{\text{\textcolor{compactor}{Gradient flow only through the }$\mathbf{z^{\scriptscriptstyle L}}_i$}}
\end{array}
\label{eq:jz_final}
\end{equation}
\end{flushleft}
\normalsize

This formulation of $L_{O\textit{g}AE}$ allows separating the influence of the $\mathbf{z^{\scriptscriptstyle SVM}}_i$ and the $\mathbf{z^{\scriptscriptstyle L}}_i$. We argue that the term weighted by $\beta_1$, which gradient flows through the $\mathbf{z^{\scriptscriptstyle SVM}}_i$, influences the frontier of the SVM, as it moves samples in directions such that it includes the misclassified $\mathbf{z^{\scriptscriptstyle L}}_i$ in the frontier: we call this term the \textcolor{expander}{\textit{expander}}. The term weighted by $\beta_2$, which gradient flows through the $\mathbf{z^{\scriptscriptstyle L}}_i$, will influence the misclassified $\mathbf{z^{\scriptscriptstyle L}}_i$, as it moves the samples in directions such that they enter the boundary drawn by the $\mathbf{z^{\scriptscriptstyle SVM}}_i$: we call this term the \textcolor{compactor}{\textit{compactor}}.

\subsection{Algorithm and implementation details}
\label{sec:method:algo_and_implem}
 The whole training procedure can be performed in two different manners. The first one consists in training first the auto-encoder with guidance from the OCSVM loss term\deleted{, based on batches of normal samples $\mathbf{x}$,} followed by a final OCSVM-training on all encoded normal data $\mathbf{z}$. While effectively being in two parts, the final OCSVM training is fairly quick and computationally inexpensive. The second manner is to use the already computed OCSVMs of the last $M$ iterations and perform a mean of the $M$ decisions functions. This second technique provides a one-step training of the model. \deleted{In the reminder of the manuscript, we use the first technique.}\revision{We use the second technique for Experiment \textcolor{nblue}{\hyperref[sec:1]{1}} and the first for Experiment \textcolor{nblue}{\hyperref[sec:2]{2}}.}

The whole procedure is summarized in the algorithm presented in the supplementary material S\ref{suppmat:algo} and technical details given in S\ref{suppmat:og_model_technical_details}.
Input batches can be of any type, including non-image if the the autoencoder is adapted. In this paper, we focus on anomaly detection in images and consider either $\mathbf{x}$ as a whole image (Experiment \textcolor{nblue}{\hyperref[sec:1]{1}}), or as an image patch (Experiment \textcolor{nblue}{\hyperref[sec:2]{2}}). When considering 2D image patches, at inference, as in \cite{alaverdyan_regularized_2020} and \cite{pinon_one-class_2023}, the central pixel of each patch is associated to a latent representation and then an anomaly score. \deleted{In this paper, t}\revision{T}he anomaly score is computed as the signed distance of the latent representation to the estimated OCSVM support. Then, a whole 2D anomaly map can be obtained by moving this patch in increments of 1 in all directions across the entire image and calculating the score of the central pixel for each position. A 3D score map can obtained by concatenating the 2D anomaly maps (see first row of Figure \ref{fig:visu_brats16} for examples of anomaly score maps superposed with MRIs).

\section{\textcolor{nblue}{Experiment 1}: Digit distinction under corruptions}
\label{sec:1}

\begin{figure}
    \centering
    \includegraphics[width=0.6\linewidth]{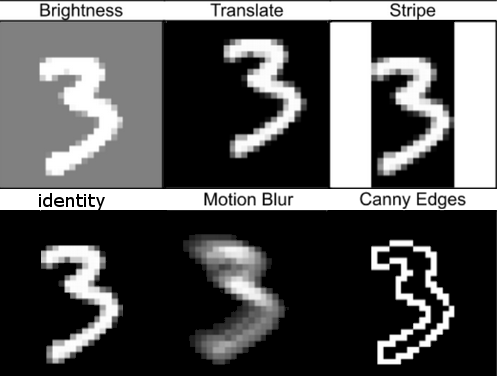}
    \caption{Corruptions of the MNIST dataset (MNIST-C \cite{MNIST-C_mu2019mnist}) used throughout this article, on the digit 3.}
    \label{fig:mnist-c}
\end{figure}

We propose  a first use-case experiment to evaluate the performance of the proposed model in a controlled setting against state-of-the-art models. The proposed task  is to evaluate if the models can correctly classify handwritten digits of the normal class versus digits of other classes when presented with a wide variety of corruption noises.

\subsection{Experimental setup and dataset}
\label{sec:1:setup}

\subsubsection{Corrupted MNIST database}
\label{sec:1:setup:mnist-c}

MNIST-C \cite{MNIST-C_mu2019mnist} is a corrupted variant of the MNIST dataset, designed to evaluate model robustness under distribution shifts. It applies 15 different types of corruptions, such as noise, blur, and geometric transformations, to the original MNIST digits images of dimension 28x28. 
%We evaluate the capability of the different models to perform anomaly detection under distribution shift, meaning that 
In this experiment, we train the networks on a ``normal" digit class corresponding here to digit \textit{3}, under specific corruptions, here \textit{identity}, \textit{motion blur} and \textit{translate}, and then evaluate the networks ability to distinguish ``normal" from anomalous digits, corresponding to digit \textit{8}, under another distribution of corruptions, here \textit{stripe}, \textit{canny edges} and \textit{brightness}, as exemplified on Figure \ref{fig:mnist-c}. The training set is composed of 18393 images, concatenating 6131 handwritten \textit{3} training images of each corruptions type, (\textit{identity}, \textit{motion blur} and \textit{translate}). 90\% of these images of the normal class are used for model training and 10\% are used for early stopping. The validation set is composed of both 974 handwritten \textit{8} and 1010 \textit{3} images from the testing set of MNIST, corrupted with the testing corruptions (\textit{stripe}, \textit{canny edges} and \textit{brightness} in Figure \ref{fig:mnist-c}), for a total of 5952 images. The testing set is composed of both 5851 handwritten \textit{8} and 6131 \textit{3} images from the training set of MNIST, different from those of the training dataset and corrupted with the testing corruptions (\textit{stripe}, \textit{canny edges} and \textit{brightness}), for a total of 35946 images. Using the original testing set for validation and the
original training set for testing allows to give the testing set the
most samples and thus the most statistical power. Note that the validation set and the testing set have no samples in common. The digit \textit{8} was chosen to resemble digit \textit{3} and thus correspond to a challenging classification task.

\subsubsection{Compared methods}
\label{sec:1:setup:compared_method}

To evaluate our proposed method, we benchmark it against a set of commonly used approaches in UAD that align with the two main paradigms discussed in Section~\ref{sec:biblio}: reconstruction-based methods (Section~\ref{sec:biblio:ae_recons}) and support estimation-based methods (Section~\ref{sec:biblio:supp_densi_methods}), both using autoencoders for representation learning.
First, we include standard Autoencoder (AE), Variational Autoencoder (VAE) and Siamese Autoencoder (SAE) models, assessing their anomaly detection performance based on reconstruction error. These models are widely used as baseline approaches in anomaly detection, as discussed in \cite{ruff_unifying_2021} and \cite{baur2021autoencoders}. Additionally, we evaluate their combination (non-coupled) with a OCSVM trained on the learned latent space, following prior works \cite{mabu2021anomaly, pinon_detection_2023, alaverdyan_regularized_2020}.
Second, we include comparison with the standard coupled methods described in (Section~\ref{sec:biblio:supp_densi_methods:coupled}), namely Deep SVDD (both \textit{hard} and \textit{soft} versions) \cite{ruff2018deep}, Deep Structure Preservation SVDD (DSPSVDD) \cite{zhang_anomaly_2021}, and Deep VAE-SVDD (DVAESVDD) \cite{zhou2021vae}

For each method, we benchmark a set of their corresponding hyperparameters, and choose the best performing hyperparameter on the validation set. Performances are then reported for the testing set. The benchmarked hyperparameters are  detailed in supplementary materials S\ref{suppmat:hp_xp1}. 

\subsubsection{Proposed task}
\label{sec:1:setup:task}

We propose to perform a classification task at the image level. The different models take as input the whole image of dimension 28x28. They either directly output an anomaly score or output an anomaly score map of the same dimension as the input image (for reconstruction methods), that is then averaged to obtain a final score.
%We train each model on images of ``normal" digits, here \textit{3}, under specific corruptions, here \textit{identity}, \textit{motion blur} and \textit{translate}, and then evaluate its ability to distinguish ``normal" from anomalous digit (here \textit{8}) at inference, under another distribution of corruptions, here \textit{stripe}, \textit{canny edges} and \textit{brightness}. 

\subsubsection{Metrics and statistical testing}
\label{sec:1:setup:metrics}

We evaluate anomaly detection performance at the image level using standard metrics including the areas under the ROC curve (\textit{AUROC}) and \textit{AUROC30} which focuses on the low false-positive rate regime ($\leq$30\%), as well as the Area Under the Precision-Recall Curve (\textit{AUPR}), which is particularly relevant for highly imbalanced datasets. 
We perform statistical testing among the compared models, by generating 1000 bootstrap samples by sampling the testing set with replacement, then computing the evaluation metrics  for each model on each bootstrap sample, and finally identifying the best-performing model based on the mean metric values. We then perform a paired bootstrap test, computing $p$-values as the fraction of bootstrap samples where a competing model outperforms the best model. To account for multiple comparisons, we apply Bonferroni correction, adjusting the significance threshold accordingly.

\subsection{Results and discussion}
\label{sec:1:results}

Table \ref{tab:1table_3v8_s1} presents the performance metrics obtained by all benchmarked models when distinguishing \textit{3} from \textit{8} under noise distribution shifts. 
We observe that performance of representation models paired with OCSVM (AE + \textit{ocsvm}, VAE + \textit{ocsvm} and SAE + \textit{ocsvm}) are on par with their reconstruction-based counterparts (AE \textit{recons. error}, VAE \textit{recons. error} and SAE \textit{recons. error}). Our proposed model, O\textit{g}AE, achieves better performances than any other model on all metrics (except when being on par with DVAESVDD for \textit{AUROC}). Overall, the basic methods (AE-based) remain competitive, consistently performing within 5 points of the best model for every metric.

\begin{table}[t]
    \centering
\caption{Performance of studied models on discriminating \textit{3} VS \textit{8} under corruption. Best model in bold. Models with no statistically significant difference (p-value $<$ 0.01 after paired bootstrap test with Bonferroni correction) are underlined.}
\label{tab:1table_3v8_s1}

    % \caption{\textbf{3 vs 8 \textit{s1}}}
    \begin{tabular}{|c|c|c|c|}
    \hline
    \textbf{3 vs 8} & \textit{AUROC} & \textit{AUPR} & \textit{AUROC30} \\ \hline
    AE \textit{recons. error} & 0.56 & 0.66 & 0.66 \\ \hline
AE + \textit{ocsvm} & 0.54 & 0.65 & 0.67 \\ \hline
VAE \textit{recons. error} & 0.54 & 0.65 & 0.66 \\ \hline
VAE + \textit{ocsvm} & 0.52 & 0.63 & 0.65 \\ \hline
SAE \textit{recons. error} & 0.55 & 0.65 & 0.67 \\ \hline
SAE + \textit{ocsvm} & 0.53 & 0.64 & 0.66 \\ \hline
O\textit{g}AE  [\textcolor{nblue}{\hyperref[sec:method]{ours}}] & \underline{0.59} & \textbf{0.70} & \textbf{0.71} \\ \hline
h-DSVDD \cite{ruff2018deep} & 0.51 & 0.62 & 0.65 \\ \hline
s-DSVDD \cite{ruff2018deep} & 0.52 & 0.63 & 0.66 \\ \hline
DSPSVDD \cite{zhang_anomaly_2021} & 0.51 & 0.62 & 0.65 \\ \hline
DVAESVDD \cite{zhou2021vae} & \textbf{0.59} & 0.67 & 0.65 \\ \hline
\end{tabular}
\end{table}

When comparing all coupled models (O\textit{g}AE, h-DSVDD, s-DSVDD, DSPSVDD and DVAESVDD), we find that O\textit{g}AE and DVAESVDD outperform their competitors  and that on this non-trivial task, some coupled models are outperformed by basic baselines (AE \textit{recons. error}), aligning with previous findings \cite{CAI2025_medianomaly}. For deep SVDD, the results consistently show that the hard-margin variant of Deep SVDD (h-SVDD) outperforms or at least matches the performance of the soft-margin version (s-SVDD). This aligns  with the original paper results \cite{ruff2018deep} and the literature,  which suggests that the added complexity of the soft-margin approach does not translate into a performance gain \cite{zhang_anomaly_2021, zhou2021vae, yi2020patch, hojjati2023dasvdd}. Additionally, we find that DVAESVDD consistently outperforms DSPSVDD. This could highlight the advantage of using the VAE for more compact latent space or adapting the center of the hyper-sphere at each batch. Both methods consistently outperform traditional Deep SVDD approaches, aligning with the findings in their original papers \cite{zhou2021vae, zhang_anomaly_2021}. 

A global analysis of the results suggests several global patterns. Models that leverage representation learning combined with explicit support estimation generally outperform or are on par with reconstruction-based methods. Coupled approaches, where representation learning and anomaly detection \deleted{are jointly optimized}\revision{influence each other}, tend to yield better results than decoupled methods. Recent methods that build upon Deep SVDD frameworks demonstrate improved performance over earlier variants. Finally, our proposed method achieves superior results on this benchmark, surpassing existing state-of-the-art models. \deleted{We found the best performing hyperparameters for our method to be $\lambda = 1^{-2}$ and $(\beta_1, \beta_2 ) = (1, 0 )$ for the first half of epochs (\textit{expander}) and $(\beta_1, \beta_2 ) = (0.5, 0.5 )$ for the last half of epochs (balanced \textit{expander} + \textit{compactor}).}\revision{By studying different \textit{compactor}/\textit{expander} ratios and strategy, we found the best configuration to be $(\beta_1, \beta_2 ) = (1, 0 )$ for the first half of epochs (\textit{expander}) and $(\beta_1, \beta_2 ) = (0.5, 0.5 )$ (balanced \textit{expander} and \textit{compactor}) for the last half of epochs. Other strategies were found to perform \textit{on par} with our selected configuration (see supplementary materials S\ref{suppmat:hp_xp1}), all of them characterized by a balanced setting of the  \textit{compactor} and \textit{expander} terms, with slightly higher values of the \textit{expander} term ($\beta_1 \gtrsim \beta_2$).} It is worth noting that the dataset corruptions introduced in our experiments can be interpreted as a form of domain shift (i.e. corruptions in the test set are not the same as those in the training set), further emphasizing the adaptability of the evaluated models in real-world scenarios.

%Parler du fait que pas de modèles compactor et ce que ça implique

\section{\textcolor{nblue}{Experiment 2}: lesion detection in brain MRI}
\label{sec:2}

In this experiment, we consider a clinically realistic scenario and evaluate the models capabilities to detect brain lesions or tumors in MRI scans. As described in Section \ref{sec:biblio:uad_medimage}, this setup is more challenging than the one in experiment \textcolor{nblue}{\hyperref[sec:1]{1}}, as brain MRI images typically contain more complex structures and noise, making anomaly detection more difficult. We consider two different applications that cover the spectrum of detection tasks mostly encountered in this domain. The first one consists in detecting and segmenting brain tumors, which are anomalies of large size but heterogeneous patterns, while the second one focuses on the detection on small and more subtle vascular lesions.

\subsection{Experimental setup and dataset}
\label{sec:2:setup}

We consider four  3D MRI T1 image databases described below, two databases of normal control subjects (\ref{sec:2:setup:control_db}), one for training and one for testing and two pathological databases for testing only (\ref{sec:2:setup:wmh}), the first one comprising exams of brain tumor patients from the public BraTS dataset \cite{Bakas_BRATS2020_ScientificData_2017} and the second one containing exams of patients with punctuate vascular lesions from the public WMH dataset \cite{kuijf_standardized_2019}. To make the setup more challenging, we only use the T1 modality for each of these databases, unlike most studies of the literature which also combine the FLAIR images, where the lesions appear as hyperintense. Also note that the training and the testing databases are totally separated, introducing domain shift, contrary to some studies which train on the healthy slices of a database and test on the pathological ones. All databases undergo the same preprocessing procedure, described in the supplementary material S\ref{suppmat:registration}, to obtain 3D volumes of size 186$\times$218$\times$135 with 1mm\textsuperscript{3} voxel size.

\subsubsection{Patient databases}
\label{sec:2:setup:wmh}

The \href{https://www.med.upenn.edu/cbica/brats2020/data.html}{\textbf{BraTS2020 dataset}} \cite{Bakas_BRATS2020_ScientificData_2017} from the 2020 brain tumor segmentation challenge consists of multi-modal brain MRI scans (T1, T1ce, T2 and FLAIR) of glioblastoma patients along with their 4-label tumor segmentation masks: background, enhancing tumor (ET), tumor core (TC) and whole tumor (WT). This dataset has also been recently employed to benchmark UAD models as described in section \ref{sec:biblio:uad_medimage}. In this analysis, we randomly selected 60 out of the 369 available T1 MRI training images of the patient cohort with a mean age of 61.2 $\pm$ 11.8 years. Examples of 2D transverse slices extracted from the 3D volume are presented in left column of Figure \ref{fig:visu_brats16}.

The \href{https://dataverse.nl/dataset.xhtml?persistentId=doi:10.34894/AECRSD}{\textbf{White Matter Hyperintensities (WMH) dataset}} originating from the WMH Segmentation Challenge \cite{kuijf_standardized_2019} consists of 60 T1w and FLAIR MRI scans from patients exhibiting small vessel diseases and acquired from three different hospitals (20 per hospital), along with expert-annotated segmentation masks. This dataset has been recently employed for the evaluation of unsupervised anomaly detection \cite{baur_modeling_2021, pinaya_unsupervised_2022, pinaya_fast_2022, meissen2021challenging, pinon_one-class_2023}, using separate normative datasets for training and leveraging WMH data exclusively for evaluation. Examples of transverse  2D T1 MRI transverse slices extracted from the 3D volume are presented in left columns of Figure \ref{fig:visu_am126}.
The patient cohort has a mean age of 70.1 $\pm$ 9.3 years  and exhibits a wide range of lesion volumes (0.78 cm³ to 195.15 cm³), making it particularly challenging due to inter-subject variability and scanner differences.

\subsubsection{Control databases}
\label{sec:2:setup:control_db}

The first normative database used for the training and validation of the different UAD models is the semi-public \textbf{CERMEP control dataset} \cite{merida_cermep-idb-mrxfdg_2021} which comprises 75 healthy controls' MRI scans. The subjects in this control group have an average age of 38 $\pm$ 11.5 years, which is younger compared to the WMH and BraTS patient cohorts. Note that this age gap may introduce a domain shift between the train and test groups potentially impacting detection performance, notably as the process of normal-brain aging results in a slight brain shrinkage.
The second control dataset used for testing is a subset of the openly available \href{https://brain-development.org/ixi-dataset/}{\textbf{IXI dataset}}, which comprises nearly 600 MRI scans from healthy subjects. For this study, we selected 60 IXI controls for testing, age-matched to the WMH dataset (70.1 ± 9.3 years), to mitigate potential age-related bias in the model's classification performance.  We considered that the age-distribution of the IXI dataset was close enough to that of the BraTS dataset (61.2 $\pm$ 11.8 years) to avoid extracting a second subset of the IXI dataset that would perfectly match the age distribution of the BraTS dataset.
%The age-matching distributions of the test normative and pathological datasets is aimed at preventing models to learn distinguishing datasets based on age rather than pathological features . to reduce computational burden and assuming that the classification and detection tasks on the BraTS dataset are less sensitive to the age-matching procedure, since the patients images exhibit larger differences due to the brain tumor presence than that observed difference between the normative and the WMH datasets where the lesions are subtle and less easily discriminated from control subjects. 

\subsubsection{Compared Methods}
\label{sec:2:setup:compared_methods}

Based on the synthetic review (section \ref{sec:biblio:uad_medimage}) of the state-of-the-art UAD approaches that have been evaluated on MRI of large and heterogeneous brain lesions, such as BraTS or more subtle lesions such as MSLUB, WMH or ATLAS (considering the small lesions only), we chose to pick one among the best performing models of each category. Considering support/density estimation methods, we selected the model combining a global siamese autoencoder and localized per-voxel OCSVM models \cite{alaverdyan_regularized_2020}, referred to as SAE + \textit{localized} OCSVM in the following. This patch-based model used an autoencoder architecture similar to the one used in Experiment \textcolor{nblue}{\hyperref[sec:1]{1}} and detailed in the supplementary material S\ref{suppmat:ae_architectures}. We include the UNET-based autoencoder (AE) architecture proposed by Baur et al. \cite{baur_modeling_2021} as well as the restoration model VQ-VAE + Transformer proposed by Pinaya et al. \cite{pinaya_unsupervised_2022} which combines a quantized autoencoder (VQ-VAE) with an autoregressive transformer in the latent space. Both AE and VQ-VAE+Transformer models process full 2D slices which are concatenated to obtain the 3D anomaly score map. Finally, we evaluate performance of the cDDPM+MHD diffusion model proposed by Behrendt et al \cite{behrendt2024mhd} which was shown to outperform DAE or anoDDPM and other masked-based models (e.g. autoDDPM \cite{bercea2023mask}) for the detection of BraTS or small stroke lesions of the ATLAS dataset.

For a fair comparison, we implement each method using the hyperparameters provided in their respective publications. Hyperparameters for our proposed method are taken to be the best performing for Experiment \textcolor{nblue}{\hyperref[sec:1]{1}} to reduce the computational burden and provide a fairer comparison to the other methods\deleted{(no hyperparameter optimization performed)}. \revision{An additional study of the hyperparameters effect on our method, for this experiment, is presented in supplementary material S\ref{suppmat:xp2_additional}.} 

For the patch-based models, (SAE + \textit{localized} OCSVM \cite{alaverdyan_regularized_2020} and our proposed model O\textit{g}AE with \textit{localized} OCSVM), we set the 2D patch size to 15$\times$15, as done in our previous work \cite{alaverdyan_regularized_2020, pinon_one-class_2023} and so as to approximately match the size of the images on Experiment \textcolor{nblue}{\hyperref[sec:1]{1}}. The 2D anomaly score maps are reconstructed as described in section \ref{sec:method:algo_and_implem}. The 3D score map is obtained by concatenating the 2D anomaly maps. For our proposed method (O\textit{g}AE), we use batch of co-localized patches, hence the name O\textit{g}AE with \textit{localized} OCSVM.

% These methods (reconstruction-based, hybrid and decoupled) have demonstrated strong performance on brain MRI anomaly detection and serve as relevant benchmarks for our evaluation, however, as already mentioned, none of them have study the performances for T1 only nor are they coupled.

\subsubsection{Proposed tasks}
\label{sec:2:setup:tasks_medical}

We consider two different tasks in this experiment, a classification task at the image-level (3D), following the evaluation protocol of the first experiment in section \ref{sec:1:setup:task}, as well as a localization (segmentation) task at the voxel-level. Both evaluations are derived from the single anomaly score map output by each model.

For the \textbf{classification} task, to obtain a single anomaly score per patient from their anomaly score map, we tested different aggregation methods (2\% percentile, mean, median, with or without ventricle removal) and found that it had little impact on the overall results. In the end, we used the 2nd percentile threshold of the anomaly scores (meaning 2\% of scores fall below this value) while excluding the ventricles\footnote{This exclusion was first motivated because ventricles tend to exhibit high anomaly scores due to age-related differences between the control and patient databases.}. 
For the \textbf{localization} task, we directly used the anomaly score maps and compared them voxel-wise to the ground-truth masks of the lesions, to obtain localization metrics.

Training of the models was done on 80\% of the CERMEP control dataset, while the remaining 20\% was used for early stopping during training. Testing was performed on the control IXI-age-matched subset, on the pathological WMH database and on the BraTS database for the classification task. For the localization task, testing was done on WMH and BraTS databases only (IXI controls indeed contain no lesions).
%As described in section \ref{sec:2:setup:control_db}, we trained two versions of each UAD, one based on the non-skull-stripped images to evaluate performance on the WMH dataset and one based on the skull-stripped images to evaluate performance on the BraTS dataset.

\label{sec:2:setup:task_localization}

%Ici il faut bien dire que dans la première partie on a comparé des méthodes sur le plan fondamental (méthodes coupled, machin machin) alors qu'ici le but c'est ok maintenant qu'on a montré notre méthode, on la compare au big SOTA sur la tâche médicale. Alors que le but d'avant c'était de benchmarké plein de méthodes d'AE, basiques (AE, VAE, recons, pas recons) et des méthodes plus coupled, dont la notre.
%On prend jz3 car meilleur en somme sur les 4 tâches

\subsubsection{Metrics and statistical testing} We use the same evaluation metrics as in Experiment \textcolor{nblue}{\hyperref[sec:1]{1}} (\textit{AUROC}, \textit{AUROC30}, and \textit{AUPR}) as detailed in section \ref{sec:1:setup:metrics}, both for the classification task (distinguishing controls from patients) and the localization task (identifying lesions within patient images). Unlike the first experiment with balanced classes, this setup introduces imbalance in the localization task, where lesion voxels are rare. \textit{AUPR} is thus critical, as it better reflects performance under imbalanced training.
For the \textbf{classification} task, we perform statistical testing following the paired bootstrap test with Bonferroni correction described in section \ref{sec:1:setup:metrics} with 1000 bootstrap samples. 
For the \textbf{localization} task, we compute each metric  per patient, thus introducing natural variability across samples (patients). We employ a Kruskal-Wallis test to detect overall differences among models, followed by Dunn’s test for pairwise comparisons with Bonferroni correction.

%\subsubsection{Post-processing}
%Cette section n'existera que si j'ajoute finalement le pp, mettre en perspective ?

\subsection{Results and discussion}
\label{sec:2:results}

\begin{table*}[t]
\centering
\captionsetup[subtable]{labelformat=simple,labelsep=colon}
\renewcommand{\thesubtable}{Table~\Roman{subtable}}

\begin{tabular}{ccc}

% =====================================================
% ==================== TOP : BraTS ====================
% =====================================================

\begin{subtable}[t]{0.25\textwidth}
\centering
            \caption*{\textcolor{white}{\textcolor{white}{Classification performance. Best model in bold. Models with no statistically significant difference (p-value $<$ 0.01 after paired bootstrap test with B}}}
\begin{tabular}{|c|}
\hline
\multirow{2}{*}{\textbf{Methods}}\\
\\ \hline
AE/UNet \cite{baur_modeling_2021}\\ \hline
VQ-VAE + Transformer \cite{pinaya_unsupervised_2022}\\ \hline
cDDPM + MHD \cite{behrendt2024mhd}\\ \hline
SAE + \textit{localized} OCSVM \cite{alaverdyan_regularized_2020}\\ \hline
O\textit{g}AE with \textit{localized} OCSVM [\textcolor{nblue}{\hyperref[sec:method]{ours}}]\\ \hline
\end{tabular}
\end{subtable}
&
\setcounter{subtable}{\value{table}}
\addtocounter{subtable}{-1}

\begin{subtable}[t]{0.32\textwidth}
\centering
\caption{Classification performance. Best model in bold. Models with no statistically significant difference (p-value $<$0.01 after paired bootstrap test with Bonferroni correction) are underlined.}
\label{tab:results_wmh_classif_new}
\begin{tabular}{|c|c|c|}
\hline
\multicolumn{3}{|c|}{\textbf{Classification IXI vs BraTS}}\\ \hline
\textit{AUROC} &
\textit{AUROC 30} &
\textit{AUPR} (0.5)\\ \hline
\textbf{1.00} &
\textbf{1.00} &
\textbf{1.00}\\ \hline
\underline{1.00} &
\underline{1.00} &
\underline{1.00}\\ \hline
\underline{1.00} &
\underline{1.00} &
\underline{1.00}\\ \hline
0.96 &
0.94 &
0.96\\ \hline
0.87 &
0.83 &
0.89\\ \hline
\end{tabular}
\end{subtable}
&
\begin{subtable}[t]{0.32\textwidth}
\centering
\caption{Localization performance. Best model in bold. Models with no statistically significant difference (p-value $<$0.01 after Kruskal-Wallis and Dunn with Bonferroni correction) are underlined.}
\label{tab:results_wmh_localization_new}
\begin{tabular}{|c|c|c|}
\hline
\multicolumn{3}{|c|}{\textbf{Localization BraTS}}\\ \hline
\textit{AUROC} &
\textit{AUROC 30} &
\textit{AUPR} (0.007)\\ \hline
0.63&0.50&0.042\\ \hline
0.75&0.64&0.088\\ \hline
\textbf{0.86}&
\textbf{0.79}&
\textbf{0.247}\\ \hline
0.55&0.59&0.081\\ \hline
0.60&0.64&0.113\\ \hline
\end{tabular}
\end{subtable}
\\
\noalign{\vskip 5pt}
% =====================================================
% ==================== BOTTOM : WMH ===================
% =====================================================

\begin{subtable}[t]{0.25\textwidth}
\centering

\begin{tabular}{|c|}
\hline
\multirow{2}{*}{\textbf{Methods}}\\
\\ \hline
AE/UNet \cite{baur_modeling_2021}\\ \hline
VQ-VAE + Transformer \cite{pinaya_unsupervised_2022}\\ \hline
cDDPM + MHD \cite{behrendt2024mhd}\\ \hline
SAE + \textit{localized} OCSVM \cite{alaverdyan_regularized_2020}\\ \hline
O\textit{g}AE with \textit{localized} OCSVM [\textcolor{nblue}{\hyperref[sec:method]{ours}}]\\ \hline
\end{tabular}
\end{subtable}
&
\hspace{3px}
\begin{subtable}[t]{0.32\textwidth}
\centering

\begin{tabular}{|c|c|c|}
\hline
\multicolumn{3}{|c|}{\textbf{Classification IXI vs WMH}}\\ \hline
\textit{AUROC}&\textit{AUROC 30}&\textit{AUPR} (0.5)\\ \hline
\underline{0.98}&\underline{0.97}&\underline{0.98}\\ \hline
\underline{0.95}&\underline{0.93}&\underline{0.96}\\ \hline
0.87&0.85&0.90\\ \hline
\textbf{0.99}&\textbf{0.98}&\textbf{0.99}\\ \hline
0.91&\underline{0.91}&\underline{0.94}\\ \hline
\end{tabular}
\end{subtable}
&
\begin{subtable}[t]{0.32\textwidth}
\centering

\begin{tabular}{|c|c|c|}
\hline
\multicolumn{3}{|c|}{\textbf{Localization WMH}}\\ \hline
\textit{AUROC}&\textit{AUROC 30}&\textit{AUPR} (0.007)\\ \hline
0.39&0.42&0.005\\ \hline
0.53&0.51&0.008\\ \hline
\textbf{0.84}&
\textbf{0.73}&
\underline{0.055}\\ \hline
0.63&0.60&0.018\\ \hline
0.61&\underline{0.71}&\textbf{0.066}\\ \hline
\end{tabular}
\end{subtable}
\end{tabular}
\end{table*}
\addtocounter{table}{1}

\begin{figure*}
    \centering
    \includegraphics[width=1\linewidth]{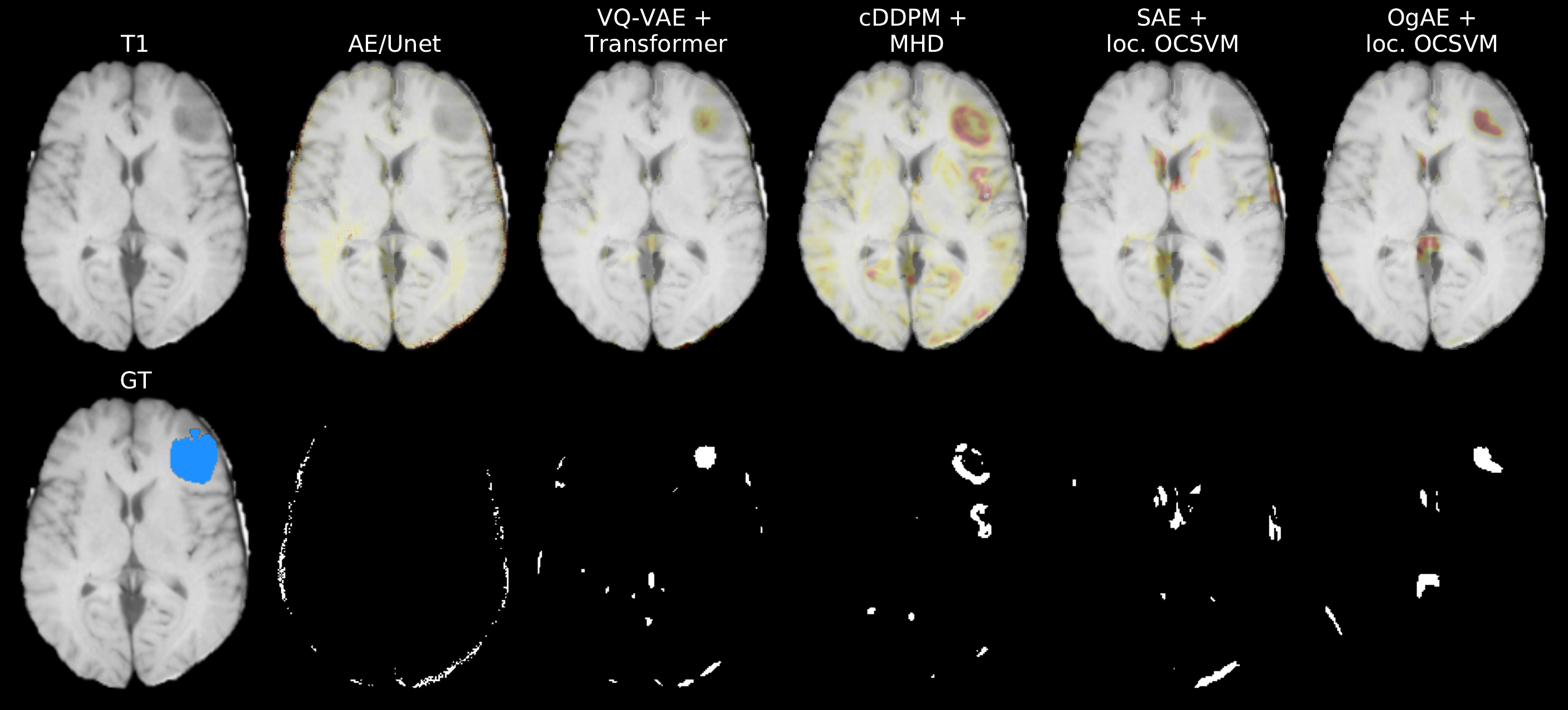}
    \caption{Visualization of a central slice from the T1-weighted brain MRI of a BraTS patient (16). The ground truth (GT) is overlaid, with light blue indicating pathological lesions. Anomaly score maps from the studied methods are superimposed, with redder colors corresponding to higher anomaly scores. At the bottom, the anomaly map is thresholded at the 2\% quantile.}
    \label{fig:visu_brats16}
\end{figure*}

\begin{figure*}
    \centering
    \includegraphics[width=1\linewidth]{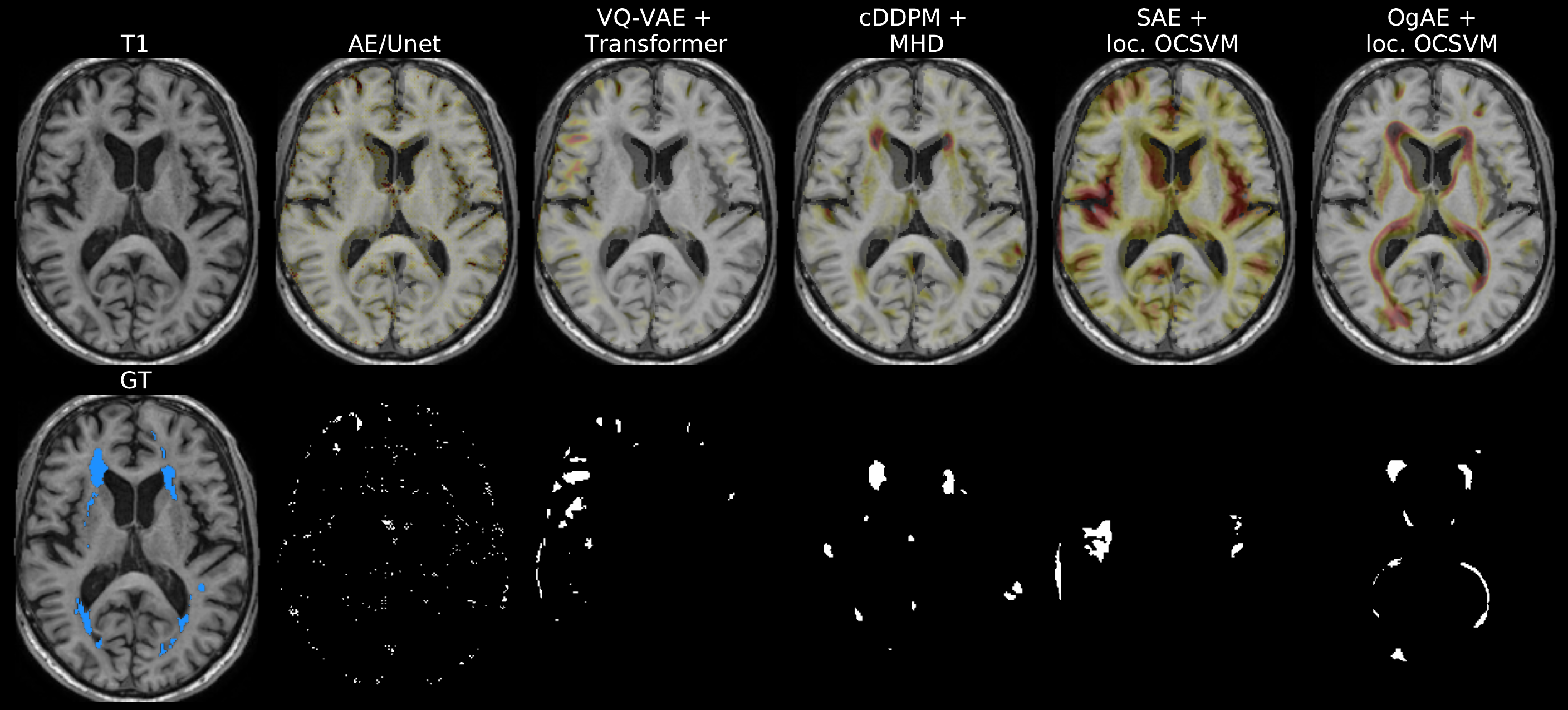}
    \caption{Visualization of a central slice from the T1-weighted brain MRI of a WMH patient (AM126). The ground truth (GT) is overlaid, with light blue indicating pathological lesions. Anomaly score maps from the studied methods are superimposed, with redder colors corresponding to higher anomaly scores. At the bottom, the anomaly map is thresholded at the 2\% quantile.}
    \label{fig:visu_am126}
\end{figure*}

Classification and localization performance on the BraTS dataset are presented in table \hyperref[tab:results_wmh_classif_new]{II} and \hyperref[tab:results_wmh_localization_new]{III} and in Figure \ref{fig:visu_brats16} (for the AE/UNet method, the dynamic range of the image had to be enhanced to [5\%, 95\%] quantile for enhanced visibility.). The three models based on reconstruction error (AE/UNet, VQ-VAE + Transformer and cDDPM + MHD) achieve perfect classification score. Our O\textit{g}AE with \textit{localized} OCSVM model achieves reasonable performance but significantly lower than the three best performing models. Performance of SAE+\textit{localized} OCSVM is lower than the best performing models. Regarding localization performance, the diffusion cDDPM + MHD model is shown to significantly outperform all other models. Our proposed O\textit{g}AE model ranks second regarding the \textit{AUPR} and \textit{AUROC30} metrics. As expected, and confirming results reported in the literature, simple AE/UNet has the lowest performance. These quantitative results are confirmed by the example score maps reported on Figure \ref{fig:visu_brats16}, highlighting the clear localization of the tumor in the anomaly score maps generated by the cDDPM + MHD (column 4) albeit balanced by a false detection, unlike our proposed O\textit{g}AE model (column 6) where the false detections are of smaller size and located close to the brain boundaries or the ventricles. Localization of these false detections may suggest that our models are likely to be slightly impacted by registration errors. These false detections, especially when close to the brain boundaries could be erased by simple post-processing. 

Classification and localization performance on the WMH dataset is presented in Tables \hyperref[tab:results_wmh_classif_new]{II} and \hyperref[tab:results_wmh_localization_new]{III} and Figure \ref{fig:visu_am126} (Dynamic range of AE/UNet had to be enhanced to [5\%, 95\%] quantile also.). All models achieve high accuracy, yet lower than performance achieved on the BraTS dataset, which can be explained by the higher difficulty of the task. We indeed emphasize that the detection and localization task on the WMH T1 MRI dataset is very challenging as exemplified on Figure \ref{fig:visu_am126} where the lesions highlighted in blue (bottom line of column 1) are small and very subtle, barely invisible on the original T1 MRI (first line of column 1). Regarding the classification task, although SAE+\textit{localized} OCSVM emerges as the best-performing method, its advantage over other approaches is not statistically significant, except regarding cDDPM + MHD which is shown to significantly underperform compared to all other models. It is important to recall that the IXI and WMH test databases are age-matched, meaning that models should not be able to distinguish images based solely on age-related degenerative changes. This ensures that any detected anomalies are not confounded by age effects. Note that the rank ordering of the different methods for this classification task is different from that achieved for classification of the BraTS dataset, with a significantly improved performance of the two support estimation models and a decrease of performance for the reconstruction-based models. Comparison of the different models based on their localization performance is slightly different, with cDDPM + MHD and our proposed model O\textit{g}AE with \textit{localized} OCSVM ranking first, with cDDPM + MHD producing the best \textit{AUROC} metric (0.84) significantly higher than that achieved by our model (0.61), while our model achieves the hightest \textit{AUPR} value of 0.066, yet not statistically better than the \textit{AUPR} achieved by cDDPM + MHD (0.055). Note that for this localization task, the baseline \textit{AUPR} (random classifier) is 0.007. AE/UNet and VQ-VAE+Transformer are not capable of localizing correctly the lesions, as their performance are at chance level or below. This result is surprising, as these models perform well at the subject level, meaning they are capable of distinguishing between control and patient, but not by directly identifying the lesions' localizations. This could suggest that these models may have found other discriminant anomalies than those annotated by the clinicians, or other  confounding features enabling discriminating  the IXI from the WMH subjects. Example visualization on Figure \ref{fig:visu_am126} is in par with the quantitative results reported in table \hyperref[tab:results_wmh_classif_new]{III}. Simple AE/UNet and VQ-VAE+Transformer do not detect the subtle WMH lesions, while our O\textit{g}AE with \textit{localized} OCSVM model produces masks best matching the ground truth lesion masks outlined in blue. cDDPM + MHD also exhibits good sensitivity to the two upper lesions but fails at detecting the thinner bottom lesions, while VQ-VAE + Transformer and SAE+\textit{localized} OCSVM estimate higher anomaly scores in cortical regions depicting a slight age-related shrinkage. This observation may be explained by the higher sensitivity of these models to the age-shift between the normative and the WMH distributions discussed in section \ref{sec:2:setup:control_db}.

In this study, we used the T1 MRI modality, where lesions are challenging to detect, unlike all previous studies performed on this database which also included MR FLAIR images where WMH lesions appear as hyperintense \cite{pinaya_fast_2022, pinaya_unsupervised_2022, baur_modeling_2021, pinon_one-class_2023, meissen2021challenging}. A broader trend emerges where models initially designed to detect hyperintense lesions struggle with this task. SAE+\textit{localized} OCSVM, originally developed for epileptogenic lesions detection (which even experts struggle to see \cite{wehner2021factors}), performs better in this context. Overall, our proposed method outperforms state-of-the-art methods or perform on par on this difficult task, particularly for identifying small lesions.

\revision{Additional \textit{expander}/\textit{compactor} strategies were investigated for our model on this experiment (see supplementary material S\ref{suppmat:xp2_additional}): we found, as in Experiment \textcolor{nblue}{\hyperref[sec:1]{1}}, that the best performing strategies correspond to a ratio of the $\beta_1/\beta_2 \gtrsim 1$.}

%Ajouter méthodes "triviales" (histo match \& co) pour dire qu'on fait pas comme certains sota de faire des méthodes de zinzin sans comparaison simple ?
% On pourrait ajouter vitef les perfs de thresholding en T1

\begin{comment}
\begin{figure*}
    \centering
    \includegraphics[width=1\linewidth]{ixi60_control6_anomaly_visualization_slice_60.pdf}   
            \caption{Visualization of a central slice from the T1-weighted brain MRI of a IXI control (IXI158-Guys-0783). Anomaly score maps from the studied methods are superimposed, with redder colors corresponding to higher anomaly scores. At the bottom, the anomaly map is thresholded at the 2\% quantile.}
    \label{fig:visu_ixi6}
\end{figure*}
\end{comment}

\section{General Discussion and conclusion}
\label{sec:discussion_conclusion}

In this work, we introduced a novel method for UAD that addresses limitations of existing approaches: most state-of-the-art methods rely either on reconstruction-based models, which have to compromise sensitivity to the anomalies and specificity impaired by the imperfect reconstruction of normal tissue, or on decoupled support/density estimation architectures where feature learning and anomaly scoring are optimized separately resulting in misaligned feature spaces. Recent attempts to couple these processes often rely on surrogate objectives, linear kernel formulations, or approximations that compromise flexibility and robustness. To overcome these challenges, we proposed a coupled framework in which the representation learning process is explicitly guided by \deleted{an analytically solvable}\revision{a} OCSVM loss that steers the encoder toward producing latent features aligned with the OCSVM decision boundary, thereby directly optimizing the feature space for anomaly detection. By enforcing this alignment during training, the encoder is encouraged to focus on features that are genuinely relevant for modeling the normative distribution, reducing overfitting to irrelevant patterns.

We evaluated our approach on two tasks: digit distinction under corruption, and  lesion detection in brain MRI. In the first task, our proposed O\textit{g}AE model outperformed both classical and state-of-the-art UAD methods, on a task  evaluating robustness to domain shifts across diverse corruptions. Early experiments where the training and testing corruptions would be the same was found too easy to discriminate the different UAD methods in this analysis (e.g. simple AE achieved \textit{AUROC} $>$ 0.83). Also, the setting where each model must distinguish between uncorrupted and corrupted digits is also fairly easy, with basic methods such as autoencoder reconstruction error reaching near perfect accuracy \cite{ruff_unifying_2021}. Additionally some corruptions  were found to naturally project to the same latent space locations, thereby making the density/support estimation trivial and the reconstructions naturally erase the corruptions, thus, to provide a challenging setup, the corruptions used have been selected such that when training a basic autoencoder, they would each be separated in its latent space, which we verified using UMAP \cite{mcinnes_umap_2020}.

In the medical imaging task, we considered two challenging detection and localization tasks in T1 MRI, one on large and heterogeneous cancer lesions from the BraTS datasets and the second one regarding the highly difficult task of detecting small and subtle WMH lesions (barely invisible to naked eye). On the BraTS dataset, comparison with state-of-the art UAD models of different categories showed that our proposed O\textit{g}AE model outperformed or matched existing models, with the exception of cDDPM + MHD.
%our proposed O\textit{g}AE model performs in par with the best performing cDDPM + MHD diffusion models for the detection of BraTS lesions,
For the detection of small subtle WMH lesions, our proposed model outperformed the compared models, as confirmed by the improved \textit{AUPR} metric. These results demonstrate, for the first time, the efficiency of coupled support estimation models compared to the reconstruction-based error models for a highly challenging clinical detection task. Additionally, at fixed hardware, our proposed method was found to train $\sim$2$\times$ faster than cDDPM + MHD and infer $\sim$100$\times$ faster. Additional computational resource analysis are performed in Supplementary material S\ref{suppmat:setup_computation}.
%Training time our method VS Behrendt : 7.5h VS 13h (\~2x faster)
%Inference time (120 volumes) our method VS Behrendt : 3.20min (0.05h) VS 5h (\~100x faster)

A key contribution of our work is the OCSVM-guided representation learning, which addresses the limitations of existing coupled approaches: it avoids the pitfalls of traditional deep SVDD approaches, which often suffer from hypersphere collapse, by ensuring that the learned representations maintain sufficient variance while still being well-clustered within the normal class. In deep SVDD, soft-margin methods explicitly model the dual-space projection through a neural network, reducing expressivity, also, the widely used hard-margin variant focuses on compacting points around a predefined center without a notion of radius. Our approach, in contrast, does not rely on a neural network projection, preserving the full expressivity of the original OCSVM formulation. Furthermore, unlike methods that arbitrarily steer all points toward a center, our model allows them to remain in place if they lie within the estimated boundary, ensuring a sufficient level of variance in the learned representation. Additionally, we think computing the loss on a holdout portion of each batch can enhance generalization.

\begin{figure}
    \centering
    \includegraphics[width=1.0\linewidth]{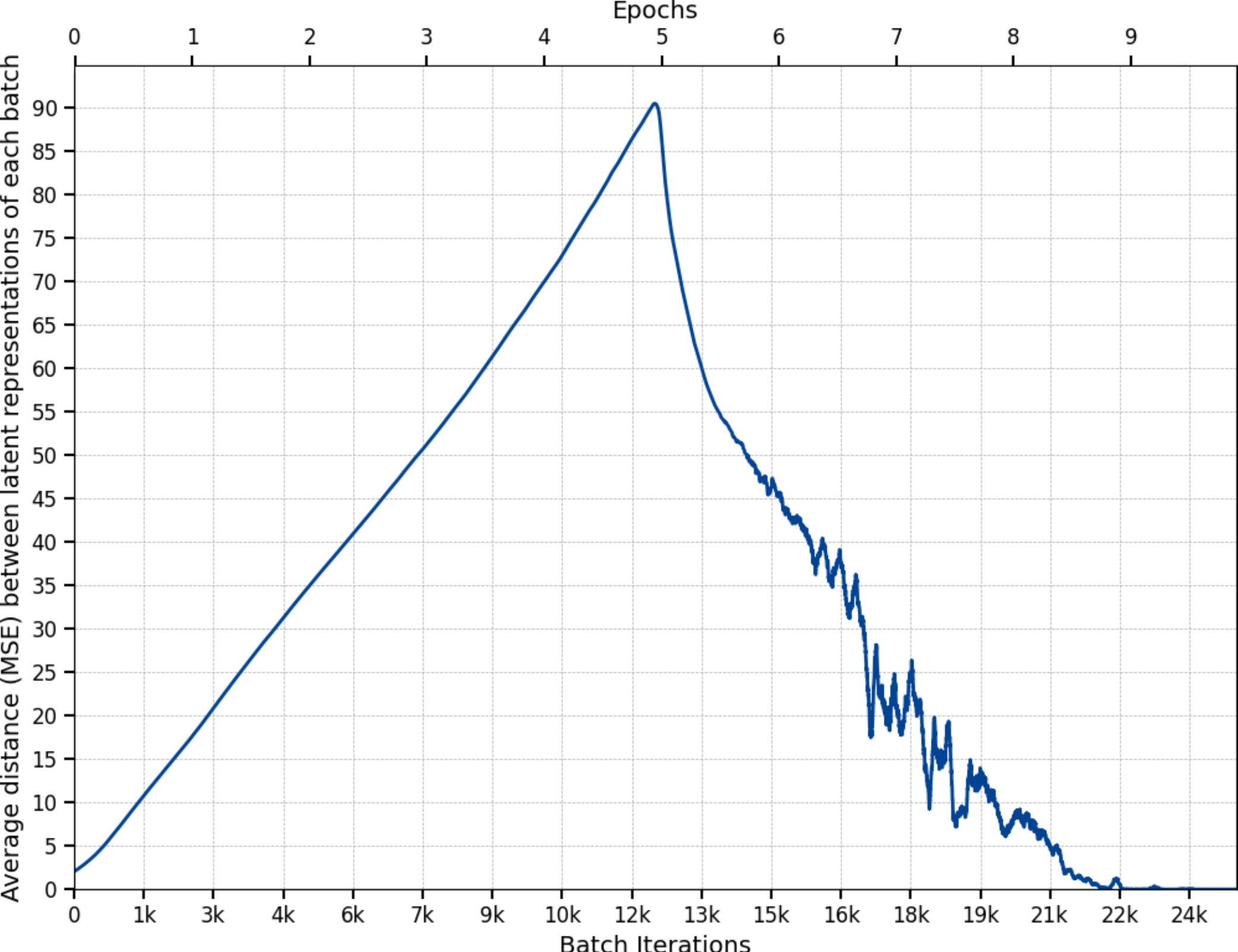}
    \caption{Average MSE between latent representations during the training of an O\textit{g}AE model for Experiment \textcolor{nblue}{\hyperref[sec:2]{2}}. For the first 5 epochs the \textcolor{expander}{expander} term is used, followed by the \textcolor{compactor}{compactor} term.}
    \label{fig:mse_z}
\end{figure}

We show, on Figure \ref{fig:mse_z}, an example of training with the \textit{expander} term (equation \ref{eq:jz_final} $\beta_1=1, \beta_2=0 $) for the first 5 epochs followed by the \textit{compactor} term ($\beta_1=0$, $\beta_2=1$) for 5 other epochs. We study the average pairwise MSE between the latent representations, which is an indicator of their spread. We see that during the expanding phase the spread of the latent representation is growing and that in the compaction phase it is decreasing, proving what we intuited. The best performing strategy \deleted{(evaluated on Experiment 1)}\revision{(on both experiments)} was found to be expander term first followed by expander + compactor with the same weight, aligning with the intuition that increasing the representation's variety at first benefits learning, but ultimately, the boundary size must be controlled and fixed. \deleted{We believe further research is needed to explore optimal training strategies.}

% SI POSSIBLE, introduire une petite discussion sur la construction du dataset a partir des données C-MNIST. ici texte dépacé de la section 4.A.3 : The setting where the training and testing corruptions would be the same was found too easy to discriminate the different UAD methods in this analysis. The setting where the method must distinguish between uncorrupted and corrupted digits is also fairly easy, with basic methods such as autoencoder reconstruction error reaching near perfect accuracy \cite{ruff_unifying_2021}. We propose the outlier digit \textit{8}, because it can be very similar to a \textit{3}, and thus is supposed to offer a more challenging setup. Also, some corruptions  were found to naturally project to the same latent space locations, thereby making the density/support estimation trivial and the reconstructions naturally erase the corruptions. To provide a difficult setup for both kind of methods, the corruptions used in the experiments have been selected such that when training a basic autoencoder, they would each be separated in its latent space, which we verified using UMAP.

For the medical image experiment, we did not employ any post-processing for our approach, unlike other works (e.g. AE/UNet in \cite{baur_modeling_2021}, VQ-VAE + Transformer in \cite{pinaya_unsupervised_2022}, cDDPM + MHD in \cite{behrendt2024mhd}), suggesting that further refinement could improve performances, particularly in the localization task. Additionally, transitioning to 3D representations for medical images could enhance the model's spatial awareness. Previous research \cite{pinon_one-class_2023} suggests that patch size has minimal impact on performance, reinforcing the generalizability of our approach. Given that the SAE+\textit{localized} OCSVM method \cite{alaverdyan_regularized_2020} was effective for epilepsy detection, we should evaluate the potential of our proposed O\textit{g}AE on epilepsy datasets as well, e.g. \cite{schuch2023open}.

Several avenues for future research remain open. While our study focused on autoencoders, the OCSVM-guided framework could be applied to other feature extraction methods (e.g. transformers). Additionally, since we have focused our study on support estimation models (see section \ref{sec:biblio:supp_densi_methods}), exploring density estimation techniques, which have proven competitive in anomaly detection, could provide further insights. Our method was designed for UAD (training only on normal samples), but in a semi-supervised setting, it could be extended by incorporating anomalous samples to refine the decision boundary: instead of only enforcing that normal samples remain inside the estimated boundary, anomalous samples could be explicitly pushed outside (or the frontier compacted such the sample remain outside). Also, an SVDD-guided variant could be implemented and evaluated, despite being similar when using the RBF kernel.

%%%%%%% ALERTE IL FAUT METTRE LES ACKNOWLEDGMENTS DANS LA VERSION FINALE %%%%%%%%%

%\section*{Acknowledgment}

%This work was funded, in whole or in part, by the \textit{Agence Nationale de la Recherche} (ANR) under projects ANR-24-CE45-4399 (SEIZURE), ANR-11-INBS-0006 (FLI) and 11-LABX-0063 (Labex PRIMES) as well as by the Fédération d’Informatique Lyonnaise (FIL), DAIAA project. This work was granted access to the HPC resources of IDRIS under the allocations 2023-AD011012813R2 made by GENCI.

% Une idée pour justifier que l'OgAE produit un meilleur espace serait de comparer sur un jeu de donnée test (ou train jsp) les valeurs propres pour voir si tous les axes sont mieux exploités

\bibliographystyle{IEEEtran}
\bibliography{bibliography_nico}

\cleardoublepage
\appendices
\section*{Supplementary material}
\pdfbookmark[1]{Supplementary material}{suppmat}
\FloatBarrier
\setcounter{subsection}{0}

\subsection{Algorithm}
\label{suppmat:algo}

We present two ways of implementing our proposed O\textit{g}AE model, one with a \textcolor{compactor}{final OCSVM training} and another with a \textcolor{expander}{storage of the last $M$ OCSVMs}. \deleted{Both have proven to be similar in terms of performances, lightweight and fast, with as few as $M=10$.}

\begin{algorithm}[H]
\caption*{Autoencoder training with OCSVM-guidance}
\label{algo}
\textbf{Input: } Normal samples $(\mathbf{x}_i)_{1 \leq i \leq N}$   \\
\textbf{Output: } Trained encoder $E$  \Comment{Decoder $D$ is discarded}
\begin{algorithmic}
\For{each epoch}
    \For{every batch of samples $(\mathbf{x}_i)_{1 \leq i \leq b}$} \Comment{Batch size $b$}
   \State Compute latent representations of samples :
   \State $(\mathbf{z}_i)_{1 \leq i \leq b} = E[(\mathbf{x}_i)_{1 \leq i \leq b}]$
   \State Split in two the $\mathbf{z}_i$ to obtain $\mathbf{z^{\scriptscriptstyle SVM}}_i$ and $\mathbf{z^{\scriptscriptstyle L}}_i$
   \State Solve the OCSVM problem for the $\mathbf{z^{\scriptscriptstyle SVM}}_i$ to obtain:
   \State $(\alpha^*_j)_{1 \leq j \leq \frac{b}{2}}$ and $\rho^*$
   \State Compute the reconstructions of latent representations:
   \State $(\mathbf{\hat{x}}_i)_{1 \leq i \leq b} = D[(\mathbf{z}_i)_{1 \leq i \leq b}]$
   \State Compute the loss (\ref{eq:jz_final}) and apply a  gradient step to $E$ and $D$
   \State \If{\textcolor{ expander}{iteration $\in$ $M$ last iterations}}
   \State \textcolor{expander}{Save $(\alpha^*_j)_{1 \leq j \leq \frac{b}{2}}$ and $\rho^*$}
   \EndIf
    \EndFor
\EndFor
\end{algorithmic}
\end{algorithm}
\vspace{-15px}
\begin{algorithm}[H]
%\ContinuedFloat
%\addtocounter{algorithm}{-1}
\caption*{\textcolor{compactor}{OCSVM final training}}
\textcolor{compactor}{\textbf{Input: } Normal samples $(\mathbf{x}_i)_{1 \leq i \leq N}$ and trained encoder $E$ \\
\textbf{Output: } Decision function $f$ of OCSVM}
\begin{algorithmic}
    \State \textcolor{compactor}{Compute latent representations of samples:}
    \State \textcolor{compactor}{$(\mathbf{z}_i)_{1 \leq N} = \mathbf{D}[(\mathbf{x}^h_i))_{1 \leq N}]$}
    \State \textcolor{compactor}{Solve the OCSVM problem for the $(\mathbf{z}_i)_{1 \leq N}$ to obtain the parameters of the final decision function}
%\For{each voxel localization $i$}
%    \State Encode the patches: $(\mathbf{z}^h_i = \mathbf{D}(\mathbf{x}^h_i))_{1 \leq h \leq N_{\mathrm{H}}}$
 %   \State Train the one-class SVM with the $N_{\mathrm{H}}$ latent representations $(\mathbf{z}^h_i)_{1 \leq h \leq N_{\mathrm{H}}}$
%\EndFor
\end{algorithmic}
\end{algorithm}

In the case of \textcolor{compactor}{final OCSVM training}, the final decision (and the encoder) is readily available for inference. In the case of \textcolor{expander}{storage of the last $M$ OCSVMs}, the mean of the $M$ decisions functions is performed at inference. \revision{For Experiment \textcolor{nblue}{\hyperref[sec:1]{1}} the relative loss of performance when choosing \textcolor{expander}{last $M$ OCSVMs} with $M=10$ over \textcolor{compactor}{final OCSVM training} was only \textit{AUROC} -0.005; \textit{AUPR} -0.01; \textit{AUROC 30} -0.005. Even though the final OCSVM training is quite inexpensive (see S\ref{suppmat:setup_computation}), the relative gain of an additional training is small.}

\subsection{Technical details for the OCSVM-guidance model}
\label{suppmat:og_model_technical_details}

This section outlines the technical implementation of the OCSVM-guidance model, particularly the gradient computation through the dual solution, the numerical stabilization techniques, and the kernel matrix reformulation.

When computing the expander term in equation \ref{eq:jz_final}, we have to differentiate through $\boldsymbol{\alpha}^*$ and $\rho^*$, thus through a convex optimization problem (the dual problem): to do this we use \cite{cvxpylayers2019}. For solver-related manner, the dual problem has to be written in a way that it is linear in parameters, not quadratic. We thus utilize the fact that $\boldsymbol{\mathbf{K}}$ is positive semi-definite (because it is a gram matrix), to express it as: $\boldsymbol{\mathbf{K}} = \boldsymbol{\mathbf{K}}^{\frac{1}{2}^T} \boldsymbol{\mathbf{K}}^\frac{1}{2}$. Where $K_{i j} = k(\mathbf{z}_i,\mathbf{z}_j)$. As recommended in \cite{chang_libsvm_2011}, because $\frac{1}{\nu_j n}$ can get very small as $n$ increase, this only leaves a tight bound for the constraint $0 \leq \alpha_{ji} \leq \frac{1}{\nu_j n}$. Thus, for numerical stability reasons, we solve a scaled problem of variable $\Tilde{\alpha}_{ji} = n \nu_j \alpha_{ji}$. As also recommended in \cite{chang_libsvm_2011} for numerical stability, to compute $\rho$, we average the $\rho$ value obtained for every support vector. Finally, also for numerical stability, we computed $\boldsymbol{\mathbf{K}}$ as $\boldsymbol{\mathbf{K}} + 1e^{-8}\boldsymbol{\mathbf{I}}$. We used the OSQP solver \cite{stellato2020osqp}.

\subsection{Benchmarked hyperparameters for Experiment \textcolor{nblue}{\hyperref[sec:1]{1}}}
\label{suppmat:hp_xp1}

\begin{itemize}
    \item Weight for KL divergence (VAE and DVAESVDD): $\beta_{\text{KL}} \in \{1, 10^{-1}, 10^{-2}\}$
    \item Weight for cosine similarity (SAE): $\alpha \in \{1, 10^{-1}, 10^{-2}\}$
    \item $\gamma$ (DSPSVDD, see article \cite{zhang_anomaly_2021}): $\gamma \in \{10^{-1}, 1, 10\}$ (balance coefficient)
    \item $\alpha$ (DVAESVDD, see article \cite{zhou2021vae}): $\alpha \in \{10^{-1}, 1, 10\}$ (balance coefficient)
    \item $\lambda$ (O\textit{g}AE): $\lambda \in \{1, 10^{-1}, 10^{-2}, 10^{-3}\}$
    \item $(\beta_1, \beta_2 )$, i.e. \textit{expander}/\textit{compactor} strategy (O\textit{g}AE) : \deleted{$(\beta_1, \beta_2 ) \in \{(1, 0 ), (0, 1), (0.5, 0.5 ) \}$ or $(0, 1)$ for the first half epoch followed by $ (0.5, 0.5 )$ or $(0, 1)$ for the first half epoch followed by $ (1, 0)$}\revision{$\beta_1 \in \{0, 0.1, 0.2, ..., 0.9, 1\}$ ($\beta_2=1-\beta_1$), with three configurations : 1) $\beta_1 $ fixed for all epochs, 2) $\beta_1 =1$ for the first half epochs than $\beta_1$ set to the evaluated value of the range \{0, 0.1, ..., 0.9, 1\} or 3) $\beta_1 $ set in the range \{0, 0.1, ..., 0.9, 1\} for the first half epochs and then $ (\beta_1, \beta_2 )=(0.5, 0.5 )$}
    \item $\nu$ (OCSVM): $\nu \in \{0.01, 0.03, 0.1, 0.3, 0.5\}$
    \item $\gamma_{\text{RBF}}$ (OCSVM): $\gamma_{\text{RBF}} \in \{10^{-1}, 10^{-2}, 10^{-3}\}$
    \item Scaling of latent variables (OCSVM): True or False
\end{itemize}

\revision{The best performing hyperparameters are presented below. For O\textit{g}AE three best performing strategies were found, two of them corresponding to considering the expander term only ($\beta_1=1$) during the first half epochs then setting $\beta_1$ to either 0.5 or 0.6 during the second part of the training  or setting ($\beta_1=0.7$) throughout the whole training.}

\begin{table}[H]
\setlength{\tabcolsep}{4pt}
\centering
\footnotesize
\begin{tabular}{|l l l c c c|}
\hline
\revision{
Method} & \multicolumn{2}{l}{\revision{Method-specific HP (best)}} & $\nu$ & $\gamma_{\text{RBF}}$ & scaler \\
\hline
\makecell[l]{AE \textit{recons.}\\\textit{error}}  & \multicolumn{2}{l}{--}                          & --   & --        & -- \\
AE + \textit{ocsvm}                                & \multicolumn{2}{l}{--}                          & 0.5  & $10^{-1}$ & T \\
\makecell[l]{VAE \textit{recons.}\\\textit{error}} & \multicolumn{2}{l}{$\beta_{\text{KL}}=10^{-2}$} & --   & --        & -- \\
VAE + \textit{ocsvm}                               & \multicolumn{2}{l}{$\beta_{\text{KL}}=10^{-1}$} & 0.1  & $10^{-3}$ & F \\
\makecell[l]{SAE \textit{recons.}\\\textit{error}} & \multicolumn{2}{l}{$\alpha=1$}                  & --   & --        & -- \\
SAE + \textit{ocsvm}\rule[-4pt]{0pt}{4pt}          & \multicolumn{2}{l}{$\alpha=10^{-2}$}            & 0.5  & $10^{-1}$ & T \\
\multirow{3}{*}{O\textit{g}AE [\textcolor{nblue}{\hyperref[sec:method]{ours}}]}
  & \multirow{3}{*}{$\lambda=10^{-2}$} & \makecell[l]{$(\beta_1,\beta_2){=}(1,0)$\\${\to}(0.5,0.5)$} & \multirow{3}{*}{0.03} & \multirow{3}{*}{$10^{-2}$} & \multirow{3}{*}{F} \\
  &                                    & \makecell[l]{$(\beta_1,\beta_2){=}(1,0)$\\${\to}(0.6,0.4)$} & \rule[-5pt]{0pt}{5pt} & & \\
  &                                    & $(\beta_1,\beta_2){=}(0.7,0.3)$\rule[-4pt]{0pt}{4pt}        & & & \\
h-DSVDD \cite{ruff2018deep}       & \multicolumn{2}{l}{--}                                       & 0.03 & $10^{-2}$ & F \\
s-DSVDD \cite{ruff2018deep}       & \multicolumn{2}{l}{--}                                       & 0.03 & $10^{-2}$ & F \\
DSPSVDD \cite{zhang_anomaly_2021} & \multicolumn{2}{l}{$\gamma=10^{-1}$}                         & 0.03 & $10^{-2}$ & F \\
DVAESVDD \cite{zhou2021vae}       & \multicolumn{2}{l}{$\alpha=1$, $\beta_{\text{KL}}=10^{-1}$}  & 0.03 & $10^{-2}$ & F \\
\hline

\end{tabular}
\end{table}

The same autoencoder and training procedure are used for every method, to ensure fair comparison. The architecture is the one used for the MNIST experiment in DVAESVDD \cite{zhou2021vae}, it is detailed in the supplementary material S\ref{suppmat:ae_architecture_xp1}. 

\begin{comment}
    
\begin{figure}
    \centering
    \includegraphics[width=0.6\linewidth]{mnist-c_modif_setup1_only.png}
    \caption{Corruptions of the MNIST dataset (MNIST-C \cite{MNIST-C_mu2019mnist}) used throughout this article, on the digit 3.}
    \label{fig:mnist-c}
\end{figure}
\end{comment}

\subsection{\revision{Additional study for Experiment \textcolor{nblue}{\hyperref[sec:2]{2}}}}
\label{suppmat:xp2_additional}

\revision{We provide below results of additional experiments with varying configurations of the $(\beta_1,\beta_2)$ hyperparameters on Experiment \textcolor{nblue}{\hyperref[sec:2]{2}}, for the WMH localization task, which we consider the hardest setup. For models with too much \textit{expander} (high value of $\beta_1$), we observed that the latent vectors $\mathbf{z}$ are pushed so far apart that the kernelized distance $k(\cdot,\cdot)\approx 0$ for any pair of distinct vectors. The decision function (equation \ref{eq:ocsvm_decision_function}) then collapses to $f(\mathbf{z}) = -\rho^*$ such that the resulting anomaly maps no longer depend on the input, varying only with spatial localization (one OCSVM, hence one $\rho^*$, is trained per voxel location in Experiment \textcolor{nblue}{\hyperref[sec:2]{2}}). This shows that on a task harder than classifying corrupted digits, the balance between the \textit{expander} and \textit{compactor} terms is more delicate and can lead to collapse.}

\begin{table}[H]
\centering
\label{table:reviewer1_wmh}
\begin{tabular}{lccc}
\toprule
\revision{Configuration} & AUROC & AUROC$30$ & AUPR \\
\midrule
\shortstack[l]{OgAE+loc.OCSVM reference\\ $(\beta_1,\beta_2)=(1,0)\to(0.5,0.5)$} & 0.61 & 0.71 & 0.066 \\
\midrule
$(\beta_1,\beta_2)=(1,0)$ & \multicolumn{3}{c}{\textit{kernel collapse}} \\
$(\beta_1,\beta_2)=(0.7,0.3)$ & \multicolumn{3}{c}{\textit{kernel collapse}} \\
$(\beta_1,\beta_2)=(1,0)\to(0.6,0.4)$ & 0.63 & 0.71 & 0.050 \\
$(\beta_1,\beta_2)=(0,1)$ & 0.62 & 0.61 & 0.013 \\
\bottomrule
\end{tabular}
\end{table}

\subsection{Brain MRI registration and preprocessing pipeline for Experiment \textcolor{nblue}{\hyperref[sec:2]{2}}}
\label{suppmat:registration}

%This subsection describes the brain MRI registration pipeline, a critical preprocessing step that enables precise voxel-wise comparisons across subjects by aligning all images to a standardized anatomical space.
The brain MRI T1 preprocessing applied in this paper is based on a pipeline implemented in SPM12 and fully described in \cite{alaverdyan_regularized_2020}. It includes a critical registration step that enables precise voxel-wise comparisons across subjects by aligning all images to a standardized anatomical space.
%Preprocessing of the T1w MR images was performed based on the reference methods implemented in SPM12. T
Spatial normalization was performed using the UniSeg which includes segmentation of grey matter, white matter and cerebrospinal fluid, correction for magnetic field inhomogeneities and spatial normalization to the standard MNI brain template. In this work, we used the default parameters for normalization and a voxel size of 1 × 1 × 1 mm. The cerebellum and brain stem were excluded from the spatially normalized images. The masking image in the reference MNI space was derived from the Hammersmith maximum probability atlas. On top of that, each image was intensity-normalized with: $X_{\text{norm}} = \frac{X - \min(X)}{\max(X) - \min(X)}$.
To account for the specificity of the BraTS data which are skull-stripped, we considered two versions of the CERMEP and IXI normative datasets, one where images contain signal from the skull and one where images are skull-stripped. Also, as the downloadable BraTS images are corrected from magnetic field inhomogeneities and skull-stripped, we thus applied an affine registration step to the MNI template to complete the preprocessing.

\begin{comment}
\begin{figure}
    \centering
    \includegraphics[width=0.35\linewidth]{control_2_cermep.pdf}
    \caption{Visualization of a central slice from the T1-weighted brain MRI of a control used for training (database \cite{merida_cermep-idb-mrxfdg_2021}, with younger mean age then the test databases).}
    \label{fig:control_2_cermep}
\end{figure}

\begin{figure*}
    \centering
    \includegraphics[width=1\linewidth]{wmh_sin67_anomaly_visualization_slice_60.pdf}
    \caption{Visualization of a central slice from the T1-weighted brain MRI of a WMH patient (SIN67). The ground truth (GT) is overlaid, with light blue indicating white matter lesions (``hyperintensities" on FLAIR MRI but not on T1) and blue representing other pathologies. Anomaly score maps from the studied methods are superimposed, with redder colors corresponding to higher anomaly scores. At the bottom, the anomaly map is thresholded at the 2\% quantile.}
    \label{fig:visu_sin67}
\end{figure*}

\begin{figure*}
    \centering
    \includegraphics[width=1\linewidth]{ixi60_control3_anomaly_visualization_slice_60.pdf}
        \caption{Visualization of a central slice from the T1-weighted brain MRI of a IXI control (IXI072-HH-2324). Anomaly score maps from the studied methods are superimposed, with redder colors corresponding to higher anomaly scores. At the bottom, the anomaly map is thresholded at the 2\% quantile.}
    \label{fig:visu_ixi3}
\end{figure*}
\end{comment}

\subsection{Autoencoder architectures}
\label{suppmat:ae_architectures}

\subsubsection{Experiment \textcolor{nblue}{\hyperref[sec:1]{1}}}
\label{suppmat:ae_architecture_xp1}

The autoencoder architecture for all models of Experiment \textcolor{nblue}{\hyperref[sec:1]{1}} is the one used for the MNIST experiment in DVAESVDD \cite{zhou2021vae}. It consists of a convolutional encoder and a symmetric decoder. The encoder comprises two convolutional layers (5×5 kernels, 4 and 8 filters), each followed by batch normalization, LeakyReLU activation, and 2×2 max pooling. The latent representation is obtained via a fully connected layer of dimension 32 (meaning reduction factor of 24.5). The decoder mirrors the encoder, employing a dense layer to reshape the latent space, followed by two transposed convolutional layers (5×5 kernels, 8 and 4 filters) interleaved with batch normalization, LeakyReLU activation, and 2×2 upsampling. A final transposed convolution (5×5, 1 filter) with a sigmoid activation reconstructs the input. The model is trained with mean squared error as the reconstruction loss, optimized with Adam (learning rate: 1e-3), with a batch size of 100, for 20 epochs.

\subsubsection{Experiment \textcolor{nblue}{\hyperref[sec:2]{2}}}
\label{suppmat:ae_architecture_xp2}

%We present in this section the autoencoder architecture used in experiment \textcolor{nblue}{\hyperref[sec:2]{2}}. 
The encoder consists of four convolutional layers: a 5×5 layer with 3 filters, followed by three successive 3×3 layers with 4, 12, and 16 filters, respectively. Each convolutional layer is paired with batch normalization and GELU activation. The decoder mirrors this structure precisely. It begins with three 3×3 transposed convolutional layers with 12, 4, and 3 filters, each followed by batch normalization and GELU activation, and concludes with a 5×5 transposed convolution and a sigmoid activation. For training, we optimize the model using mean squared error (MSE) with the Adam optimizer (learning rate: 1e-3), trained for 10 epochs with a batch size of 100.

\subsection{Hardware setup and computation performances analysis}
\label{suppmat:setup_computation}

The majority of the computations were run on a local computer equiped with a GeForce GTX 1660 SUPER (6Gb VRAM), along with 16 Gb of RAM and and AMD Ryzen 5 3600 6-Core Processor. Occasionally when launching large batches of experiments, a HPC equiped with a NVIDIA Tesla V100 (32Gb VRAM) was used.

For Experiment \textcolor{nblue}{\hyperref[sec:1]{1}}, for all methods, among training of the autoencoder model, training of the final ocsvm (if there was one) and inference, the training of the autoencoder was always the longest of the three. OCSVM training time (when applicable) was below 1.37s for every method (only using CPU) and inference time was between 2.27s and 8.12s depending on the models (longest being AE and quickest DVAESVDD). Training of O\textit{g}AE was the longest time with 465.49s, with other methods being between 38.77s and 64.47s. Mean GPU power during training was between 37.0W for SAE and 61.1W for O\textit{g}AE. Mean CPU and RAM usage were very similar (14\% difference at max for RAM and 18\% at max for CPU) between each methods. We believe that these computation costs, while measured on Experiment \textcolor{nblue}{\hyperref[sec:1]{1}}, still holds for Experiment \textcolor{nblue}{\hyperref[sec:2]{2}} (for the patch-based methods), where the patches used are smaller (15x15 in Experiment \textcolor{nblue}{\hyperref[sec:2]{2}} against 28x28 in Experiment \textcolor{nblue}{\hyperref[sec:1]{1}}).

For Experiment \textcolor{nblue}{\hyperref[sec:2]{2}}, as stated in section \ref{sec:discussion_conclusion}, our proposed method was found to train $\sim$2$\times$ faster than cDDPM + MHD (7.5h VS 13h) and infer $\sim$100$\times$ faster (3.20min = 0.05h VS 5h). Training time of the AE/UNet was 4.9h and VQ-VAE + Transformer was 18.6h (10h for VQ-VAE and 8.6h for Transformer). As our proposed method is patch-based, we expect that its computational cost will only grow linearly as a function of the image size, contrary to methods that take as input full images.

\end{document}